\documentclass[10pt,journal,compsoc]{IEEEtran}

\usepackage{amsmath}
\usepackage{mathrsfs}  
\usepackage{color,soul}
\usepackage{bm}
\usepackage{dsfont}
\usepackage{graphicx}
\usepackage{multirow}
\usepackage{subcaption}
\usepackage{amssymb}
\usepackage{balance}

\usepackage{algorithm}
\usepackage{algorithmic}
\usepackage{capt-of}

\soulregister\cite7
\soulregister\citet7
\soulregister\citeA7
\soulregister\shortcite7
\soulregister\shortciteA7
\soulregister\citep7
\soulregister\ref7
\soulregister\pageref7
\soulregister\st7
\newcommand*{\QEDA}{\hfill\ensuremath{\blacksquare}}

\def\mycmd{0}

\if\mycmd1
\usepackage{tikz}
\usetikzlibrary{shapes, shapes.geometric, shapes.symbols, shapes.arrows, shapes.multipart, shapes.callouts, shapes.misc}
\usetikzlibrary{fit,positioning}
\usetikzlibrary{patterns}
\usepackage[utf8]{inputenc}
	\usepackage{fontspec}
	\usepackage{pgfplots}
	\usepgfplotslibrary{groupplots}
	\newlength\figurewidth
	\newlength\figureheight
	\pgfplotsset{compat=1.9}
	\usepgfplotslibrary{external} 
	\tikzset{external/system call={lualatex
	\tikzexternalcheckshellescape -halt-on-error -interaction=batchmode -jobname "\image" "\texsource"}}
\fi

\usepackage[numbers]{natbib}
\usepackage{soul}
\newcommand\bfu{\mathbf u}

\newcommand\bfx{\mathbf x}
\newcommand\bff{\mathbf f}
\newcommand\bfy{\mathbf y}

\newcommand\bfZ{\mathbf Z}
\newcommand\bfN{\mathbf N}
\newcommand\bfm{\mathbf m}
\newcommand\bfS{\mathbf S}
\newcommand\bfK{\mathbf K}
\newcommand\bfalpha{\bm \alpha}
\newcommand\bfgamma{\bm \gamma}
\newcommand\bfmu{\boldsymbol \mu}
\newcommand\bfSigma{\boldsymbol \Sigma}
\newcommand{\dee}{\,\textrm{d}}
\newcommand\indep{\protect\mathpalette{\protect\independenT}{\perp}}
\def\independenT#1#2{\mathrel{\rlap{$#1#2$}\mkern2mu{#1#2}}}

\newcommand{\ubar}[1]{\underline{#1}}
\newcommand{\obar}[1]{\overline{#1}}

\ifCLASSINFOpdf
\else
\fi
\begin{document}

\title{Scalable Joint Models for Reliable Uncertainty-Aware Event Prediction}

\author{Hossein~Soleimani,~\IEEEmembership{}
        James~Hensman,~\IEEEmembership{}
        and~Suchi~Saria~\IEEEmembership{}
\IEEEcompsocitemizethanks{\IEEEcompsocthanksitem H. Soleimani (E-mail: hsoleimani@jhu.edu) and S. Saria (E-mail: ssaria@cs.jhu.edu) are with the Department of Computer Science, Johns Hopkins University,
       Baltimore, MD 21218, USA.\protect
\IEEEcompsocthanksitem J. Hensman (Email: james.hensman@lancaster.ac.uk) is with the Division of Medicine, Lancaster University, Lancaster, LA1 4YB, UK.}
\thanks{Manuscript received March 3, 2017; revised June 19, 2017; accepted for publication August 15, 2017.}}

\markboth{IEEE Transactions on Pattern Analysis and Machine Intelligence}
{Shell \MakeLowercase{\textit{et al.}}: Bare Demo of IEEEtran.cls for Computer Society Journals}

\IEEEtitleabstractindextext{%
\begin{abstract}
Missing data and noisy observations pose significant challenges for reliably predicting events from irregularly sampled multivariate time series (longitudinal) data. Imputation methods, which are typically used for completing the data prior to event prediction, lack a principled mechanism to account for the uncertainty due to missingness. Alternatively, state-of-the-art \emph{joint modeling} techniques can be used for jointly modeling the longitudinal and event data and compute event probabilities conditioned on the longitudinal observations. These approaches, however, make strong parametric assumptions and do not easily scale to multivariate signals with many observations. Our proposed approach consists of several key innovations. First, we develop a flexible and scalable joint model based upon sparse multiple-output Gaussian processes. Unlike state-of-the-art joint models, the proposed model can explain highly challenging structure including non-Gaussian noise while scaling to large data. Second, we derive an optimal policy for predicting events using the distribution of the event occurrence estimated by the joint model. The derived policy trades-off the cost of a delayed detection versus incorrect assessments and abstains from making decisions when the estimated event probability does not satisfy the derived confidence criteria. Experiments on a large dataset show that the proposed framework significantly outperforms state-of-the-art techniques in event prediction.

\end{abstract}

\begin{IEEEkeywords}
  Uncertainty-Aware Prediction, Missing Data, Scalable Gaussian Processes, Survival Analysis, Joint Modeling, Time Series
\end{IEEEkeywords}}

\maketitle

\IEEEdisplaynontitleabstractindextext

%
\IEEEpeerreviewmaketitle

\bstctlcite{IEEEexample:BSTcontrol}

\begin{figure}[t]
\centering
\includegraphics[width=0.48\textwidth]{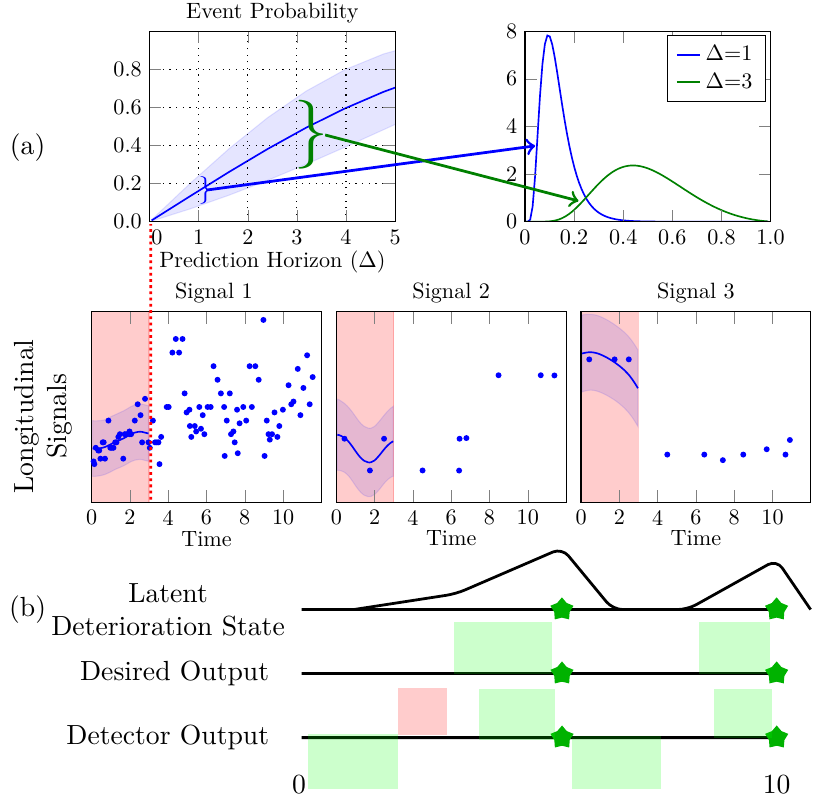}\vspace{-3mm}
\caption{(a) shows estimates from a joint-model over longitudinal and time-to-event data. Data from the shaded red region are used to estimate the probability of occurrence of the event. Further, within a given $\Delta$, the distribution of the event probability is shown in the top right. (b) describes the observed event data (green stars). The latent deterioration state shows an example pattern that may lead to the observed events. Here, the patient gradually transitions from being healthy to becoming sick and when they get worse enough, the symptoms associated with the event---in this case, septic shock---become visible. For the desired output, ideally the system should identify that the patient is deteriorating as soon as it starts to occur. For the detector output, a positive (or negative) prediction is shown as above (or below) the axis. The color indicates whether the prediction is correct (green) or wrong (red). At a given time, the detector may choose to not predict. This is shown as intervals where neither a positive nor negative prediction is made. Here, a detection much prior to the event time is considered a false detection.}
\label{main_figure}
\end{figure}

\IEEEraisesectionheading{\section{Introduction}\label{sec:introduction}}
\vspace{-1mm}
\IEEEPARstart{W}{e} are motivated by the problem of predicting events from noisy, multivariate \emph{longitudinal data}---repeated observations that are irregularly-sampled \cite{verbeke2009linear}. As an example application, consider the challenge of reliably predicting impending adverse events in the hospital. Many life-threatening adverse events such as sepsis and cardiac arrest are treatable if detected early \cite{NEJMoa010307,Schein19901388,Smith1998133,kumar2006duration}.
Towards this, one can leverage the vast number of signals---e.g., heart rate, respiratory rate, blood cell counts, creatinine---that are already recorded by clinicians over time to track an individual's health status. However, repeated observations for each signal are not recorded at regular intervals. Instead, the choice of when to record is driven by the clinician's index of suspicion. For example, if a past observation of the blood cell count suggests that the individual's health is deteriorating, they are likely to order the test more frequently leading to more frequent observations. Further, different tests may be ordered at different times leading to different patterns of missingness across different signals (see example shown in Fig. \ref{main_figure}a). Problems of similar nature arise in monitoring the health of data centers and predicting failures based on the longitudinal data of product and system usage statistics \cite{pelkonen2015gorilla}.

In statistics, the task of event prediction is cast under the framework of time-to-event or survival analysis \cite{Kalbfleisch2011,Houwelingen2012}. Here, there are two main classes of approaches. In the first, the longitudinal and event data are modeled \emph{jointly} and the conditional distribution of the event probability is obtained given the longitudinal data observed until a given time; e.g.,  \cite{Rizopoulos2012a,Rizopoulos2011a,Rizopoulos2010,Proust-Lima2014,Proust-Lima2016,rizopoulos2014combining,futoma2016}. \citet{Rizopoulos2010}, for example, posits a linear mixed-effects (LME) model for the longitudinal data. The time-to-event data are linked to the longitudinal data via the LME parameters. Thus, given past longitudinal data at any time $t$, one can compute the conditional distribution for probability of occurrence of the event within any future interval $\Delta$ (as shown in Fig. \ref{main_figure}a). \citet{futoma2016} allow a more flexible model that makes fewer parametric assumptions: specifically, they fit a mixture of Gaussian processes but they focus on single time series. In general, state-of-the-art techniques for joint-modeling of longitudinal and event data require making strong parametric assumptions about the form of the longitudinal data in order to scale to multiple signals with many observations. This need for making strong parametric assumptions limits applicability to challenging time series (such as those in our example application). An alternative class of approaches uses \emph{two-stage} modeling: features are computed from the longitudinal data and a separate time-to-event predictor is learned given the features \cite{Tsiatis2004,Sweeting2011}. For signals that are irregularly sampled, the missing values are completed using imputation and point estimates of the features are extracted from the completed data for the time-to-event model (e.g., \citet{Henry2015}). An issue with this latter class of approaches is that they have \emph{no principled means of accounting for uncertainty due to missingness}. For example, features may be estimated more reliably in regions with dense observations compared to regions with very few measurements. But by ignoring uncertainty due to missingness, the resulting event predictor is more likely to trigger false or missed detections in regions with unreliable feature estimates.

Alternatively, one can treat event forecasting as a time-series classification task. This requires transforming the event data into a sequence of binary labels, $1$ if the event is likely to occur within a given horizon and $0$ otherwise. However, to binarize the event data, we must assume a fixed horizon $(\Delta)$. Further, by doing so, we lose valuable information about the precise timing of the event (e.g., information about whether the event occurs at the beginning or near the end of the horizon $\Delta$). For prediction, a sliding window is used for computing point estimates of the features by using imputation techniques to complete the data or by using model parameters from fitting a sophisticated probabilistic model to the time-series data \cite{Ghassemi2015,Wuocw138}. These methods suffer from similar shortcomings as the two-stage time-to-event analysis approaches described above: they do not fully leverage uncertainty due to missingness in the longitudinal data. We discuss these works in more detail in Section \ref{related_work}.

In this paper, we explore the following question: can we exploit uncertainty due to missingness in the longitudinal data to improve reliability of predicting future events? We propose a \emph{reliable event prediction} framework comprising two key innovations.

\textbf{1)} We propose a flexible Bayesian nonparametric model for jointly modeling the high-dimensional, multivariate longitudinal and time-to-event data. Specifically, this model is used for computing the probability of occurrence of an event, $H(\Delta| \bfy^{0:t}, t)$, within any given horizon $(t, t+\Delta]$ conditioned on the longitudinal data $\bfy^{0:t}$ observed until $t$. Compared with existing state-of-the-art in joint modeling, the proposed approach scales to large data without making strong parametric assumptions about the form of the longitudinal data. Specifically, we relax the need to assume simple parametric models for the time series data. We use multiple-output Gaussian Processes (GPs) to model the multivariate, longitudinal data. This accounts for non-trivial correlations across the time series while flexibly capturing structure within a series. Further, in order to facilitate scalable learning and inference, we propose a stochastic variational inference algorithm that leverages sparse-GP techniques.  {This reduces the complexity of inference from cubic in the number of observations per signal ($N$) and the number of signals ($D$) to linear in both $N$ and $D$.}

\textbf{2)} We use a decision-theoretic approach to derive an optimal detector which uses the predicted event probability $H(\Delta|\bfy^{0:t}, t)$ and its associated uncertainty to trade-off the cost of a delayed detection versus the cost of making incorrect assessments. As shown in the example detector output in Fig. \ref{main_figure}b, the detector may choose to wait in order to avoid the cost of raising a false alarm. Others have explored other notions of reliable prediction. For instance, classification with abstention (or with rejection) has been studied before (see, e.g.,  \cite{chow1957optimum,golfarelli1997error,bartlett2008classification}). Decision making in these methods are based on point-estimates of the features and the event probabilities. Others have considered reliable prediction in classification of segmented video frames each containing a single class. In these approaches, the goal is to determine the class label as early as possible \cite{Parrish2013,Sangnier2016,Hoai2014}.

The rest of the paper is organized as follows. Section \ref{background_section} reviews survival analysis and joint models. In section \ref{methods_section}, we present our joint modeling framework. Then, in section \ref{robust_section}, we develop our robust prediction policy. We review related work in section \ref{related_work}. In section \ref{exp_section}, we show results on a challenging dataset from patients admitted to a hospital for the task of predicting a deadly adverse event called septic shock.  Finally, concluding remarks are in section \ref{conc_section}.

\section{Background: Survival Analysis}\label{background_section}
\label{surv_sect}
In this section, we review survival analysis and joint models.
Survival analysis is a class of statistical models developed for predicting and analyzing {\em survival time}: the remaining time until an event of interest happens. This includes, for instance, predicting time until a mechanical system fails or until a patient experiences a septic shock. The main focus of survival analysis is computing survival probability; i.e., the probability that each individual survives for a certain period of time given the information observed so far.

More formally, for each individual $i$, let $T_i \in \mathbb{R}^+$ be a non-negative continuous random variable representing the occurrence time of an impending event. In survival analysis, this random variable is usually characterized using a \textit{survival function}, $S(t) = Pr(T\geq t)$; i.e., the probability that the individual survives up to time $t$. Given the survival function, we can compute the probability density function $p(t) = -\frac{\partial}{\partial t} S(t)$. In survival analysis, this distribution is usually specified in terms of a \textit{hazard function}, $\lambda(t)$, which is defined as the instantaneous probability that the event happens conditioned on the information that the individual has survived up to time $t$; i.e.,
\begin{align}
\lambda(t) &\triangleq \lim_{\Delta\rightarrow 0} \frac{1}{\Delta}Pr(t < T \leq t+\Delta|T\geq t) \nonumber \\
&= \frac{p(t)}{S(t)} = -\frac{\partial}{\partial t} \log S(t).
\label{haz_def}
\end{align}

From (\ref{haz_def}), we can easily obtain $S(t) = \exp(-\int_0^{t}\lambda(s)\dee s)$ and $p(t)=\lambda(t)\exp(-\int_0^{t}\lambda(s)\dee s)$.

In the special case of $\lambda(t)=\lambda_0$, where $\lambda_0$ is a constant, this distribution reduces to the exponential distribution with $p(t)=\lambda_0\exp(\lambda_0 t)$.  In general, the hazard (risk) function may depend on some time-varying factors and individual-specific features.
A standard parametric choice for hazard function for an individual who has survived up to time $t$ is
\begin{align}
\lambda(s;t)= \lambda_0(s;t)\exp(\bfgamma^T \bfx_{it} + \bfalpha^T \bff^{0:t}_i)\, , \forall s\geq t\, ,
\label{haz_func}
\end{align}
where $\bff^{0:t}_i$ is a vector of features estimated based on longitudinal observations up to time $t$,  {$\bfx_{it}$ is a vector of observed time-invariant (e.g., gender) and time-varying (e.g., time since receiving antibiotics) covariates, and $\bfalpha$ and $\bfgamma$ are vectors of free parameters which are learned \cite{Kalbfleisch2011}}. Also, $\lambda_0(s;t)$ is a baseline hazard function which specifies the natural evolution of the risk for all individuals independently of the individual-specific features. Typical parametric forms for $\lambda_0(s;t)$ are piece-wise constant functions and $\lambda_0(s;t)=\exp(b+a(s-t)),\forall s\geq t$, where $a$ and $b$ are free parameters \cite{Kalbfleisch2011}. In this paper, we choose the latter form. { {We condition on the difference between $s$ and $t$ instead of $s$ because, in our application, a priori the time of prediction has no bearing on the risk of the event.}}.

Given this hazard function, a quantity of interest in time-to-event models is \textit{event probability} (failure probability), which is defined as the probability that the event happens within the next $\Delta$ hours:
\begin{align}
H(\Delta|\bff^{0:t}_i,t) &\triangleq 1 - S(t+\Delta|\bff^{0:t}_i,t) = P(T \leq t+\Delta|\bff^{0:t}_i, T\geq t) \nonumber \\
&= 1-\exp(-\int_{t}^{t+\Delta}\!\lambda(s;t)\dee s)\, ,
\label{failure_prob}
\end{align}
 {where $S(t+\Delta|\bff^{0:t}_i,t)\triangleq S(t+\Delta|\bff^{0:t})/S(t|\bff^{0:t})$}. The event probability, $H(\Delta|\bff^{0:t}_i,t)$, is an important quantity in many applications. For instance, (\ref{failure_prob}) can be used as a risk score to prioritize patients in an intensive care unit and allocate more resources to those with greater risk of experiencing an adverse health event in the next $\Delta$ hours. Such applications require dynamically updating failure probability as new observations become available over time.


\textbf{Joint Modeling:} The hazard function (\ref{haz_func}) and the event probability (\ref{failure_prob}) assume that the features $\bff^{0:t}_i$ are deterministically computed from the longitudinal data up to time $t$. However, computing these features may be challenging in the setting of longitudinal data with missingness. In this setting, probabilistic models are developed to jointly model the longitudinal and time-to-event data.

Let $\bfy_i^{0:t}$ be the longitudinal data up to time $t$ for individual $i$. The longitudinal component models the time series $\bfy_i^{0:t}$ and estimates the distribution of the features conditioned on $\bfy_i^{0:t}$; i.e., $p(\bff^{0:t}_i|\bfy_i^{0:t})$. Given this distribution, the time-to-event component models the survival data and estimates the event probability.

Note that because the features are random variables with distribution $p(\bff^{0:t}_i|\bfy_i^{0:t})$, the event probability $H(\Delta|\bff^{0:t}_i,t)$ is now a random quantity; i.e., every realization of the features drawn from $p(\bff^{0:t}_i|\bfy_i^{0:t})$ computes a different estimate of the event probability. As a result, the random variable $\bff^{0:t}_i$ induces a distribution on $H(\Delta|\bff^{0:t}_i,t)$: i.e., $p_H(H(\Delta|\bff^{0:t}_i,t)=h)$. This distribution is obtained from the distribution $p(\bff^{0:t}_i|\bfy_i^{0:t})$ using change-of-variable techniques (see, e.g., \citet{billingsley2008probability}).

Typically, expectation of $H(\Delta|\bff^{0:t}_i,t)$ is computed for event prediction:
\begin{align}
\bar{H}(\Delta,t) \triangleq \int H(\Delta|\bff^{0:t}_i,t) p(\bff^{0:t}_i|\bfy_i^{0:t}) \dee \bff^{0:t}_i = \int h p_H(h) \dee h. \vspace{-2mm}
\end{align}
However, we could also consider variance or quantiles of this distribution to quantify the uncertainty in the estimate of the event probability (see Fig. \ref{main_figure}).

\textbf{Learning:} Joint models maximize the joint likelihood of the longitudinal and time-to-event data, $\prod_{i=1}^I p(\bfy_i, T_i)$, where $p(\bfy_i, T_i) = \int p(\bfy_i|\bff_i)p(T_i|\bff_i)\dee \bff_i$. In many practical situations, the exact event time for some individuals is not observed due to \textit{censoring}. We consider two types of censoring: right censoring and interval censoring. In right censoring, we only know that the event did not happen before time $T_{ri}$ but the exact time of the event is unknown. Similarly, in interval censoring, we only know that the event happened within a time window, $T_{i}\in [T_{li}, T_{ri}]$.
Given these partial information, we write the likelihood of the time-to-event component $p(\mathbf{T}_i, {\delta}_i|\bff_i)$, with $\mathbf{T}_i=\{T_i, T_{ri}, T_{li}\}$ and
\begin{align}
p(\mathbf{T}_i, \delta_i|\bff_i) =
\begin{cases}
\lambda(T_i)S(T_i),\hspace{-2mm}& \text{if event observed } (\delta_i=0),\\
S(T_{li}), \hspace{-2mm}& \text{if right censored } (\delta_i=1),\\
S(T_{li})-S(T_{ri}),\hspace{-2mm}& \text{if interval censored } (\delta_i=2),\\
\end{cases}
\label{t2e_lkh}
\end{align}
where we dropped the explicit conditioning on $\bff_i$ in $\lambda(T_i|\bff_i)$ and $S(T_i|\bff_i)$ for brevity.

The value of the hazard function (\ref{haz_func}) for each time $s\geq t$ depends on the history of the features $\bff^{0:t}$. Alternatively, the hazard rate can be defined as a function of instantaneous features; i.e., $\lambda(s) = \lambda_0(s)\exp(\bfgamma^T\bfx_{is}+\bfalpha^T \bff_i(s)), \forall s$, \cite{Rizopoulos2012a}. The latter requires an accurate model for extrapolating the features (i.e.  computing $\bff^{t:t+\Delta}$ conditioned on $\bfy^{0:t}$) over the duration of $12-48$ hours to compute the survival probability $S(t+\Delta|\bfy^{0:t})=E_{[\bff^{0:t+\Delta}|\bfy^{0:t}]}\exp\big(-\int_0^{t+\Delta}\lambda(s)\dee s\big)$---this is challenging and therefore we do not include dependence on instantaneous features for our problem domain.

For training, we evaluate the likelihood for each individual at a series of grid points $t_{i1}\leq t_{i2} \leq ... \leq T_i$. At each grid point $t$, the likelihood is evaluated based on the longitudinal data observed up to time $t$ and the time-to-event component with survival time $T_i-t$ and hazard function $\lambda(s;t),\forall s\geq t$. The final training objective is the sum of logarithm of these likelihoods at each of the grid points \cite{Henry2015,VanHouwelingen2007}. Evaluating the objective at multiple grid points leading up to the event facilitates learning weights for the hazard function that prioritize features which are estimated from partial traces and are highly associated with the occurrence of an adverse event downstream. By contrast, the classical approach of evaluating the likelihood based on the complete longitudinal and the event data (e.g., \citet{Rizopoulos2012a}) is useful for retrospective analyses but poorly suited for the setting of early warning.

\vspace{-2mm}
\section{Joint Longitudinal and Time-to-Event Model}
\vspace{-0.5mm}
\label{methods_section}
In this section, we describe our framework to jointly model the longitudinal and time-to-event data. Our probabilistic joint model consists of two sub-models: a longitudinal sub-model and a time-to-event sub-model. Intuitively, the time-to-event model computes event probabilities conditioned on the features estimated in the longitudinal model. These two sub-models are learned together by maximizing the joint likelihood of the longitudinal and time-to-event data.

Let $\mathbf{y}^{0:t}_i$ be the observed longitudinal data for individual $i$ until time $t$. We develop a probabilistic joint modeling framework by maximizing the likelihood $\prod_{i} p(\mathbf{T}_i, \delta_i, \mathbf{y}^{0:t}_i)$,
where $T_i$ and $\delta_i$ are the time-to-event information defined in section \ref{surv_sect}. Unless there is ambiguity, we suppress superscripting with $t$ hereon.

In the rest of this section, we first introduce the two sub-models. This specifies the distribution $p(\mathbf{T}_i, \delta_i, \mathbf{y}^{0:t}_i)$. Then, we describe how we jointly learn these longitudinal and time-to-event sub-models.

\vspace{-2mm}
\subsection{Longitudinal Sub-model}
We use multiple-output Gaussian processes to model multivariate longitudinal data for each individual. GPs provide flexible priors over functions which can capture complicated patterns exhibited by clinical data. We develop our longitudinal sub-model based on the linear models of coregionalization (LMC) framework \cite{Journel1978,Teh2005,Alvarez2009}. LMC can naturally capture correlations between different signals of each individual. This provides a mechanism to estimate sparse signals based on their correlations with more densely sampled signals.

Let $\mathbf{y}_{id} = y_{id}(\mathbf{t}_{id}) = \{y_{id}(t_{idn}),\forall n=1,2,...,N_{id}\}$ be the collection of $N_{id}$ observations for signal $d$ of individual $i$. We denote the collection of observations of $D$ longitudinal signals of individual $i$ by $\mathbf{y}_i = \{\mathbf{y}_{i1},...,\mathbf{y}_{iD}\}$. We assume that the data are missing-at-random (MAR); i.e., the missingness mechanism does not depend on unobserved factors.
Under this assumption, we can ignore the process that caused missing data and infer parameters of the model only based on the observed data (see \citet{rubin1976inference} and Appendix B of \citet{schulam2016integrative} for a longer discussion).

We express each signal $y_{id}(t)$ as:
\begin{align}
y_{id}(t) &= f_{id}(t) + \epsilon_{id}(t)\, , \nonumber \\
f_{id}(t) &= \underbrace{\sum_{r=1}^R w_{idr} g_{ir}(t)}_{\substack{\text{shared component}}} + \underbrace{\kappa_{id} v_{id}(t)}_{\substack{\text{signal-specific}\\ \text{component}}}\, ,\label{model}
\end{align}
where $g_{ir}(t), \forall r=1,2,...,R,$ are shared latent functions, $v_{id}(t)$ is a signal-specific latent function, and $w_{idr}$ and $\kappa_{id}$ are, respectively, the weighting coefficients of the shared and signal-specific terms.

Each shared latent function $\mathbf{g}_{ir} = g_{ir}(\mathbf{t}_{id})$ is a draw from a GP with mean 0 and covariance  $\mathbf{K}^{(ir)}_{N_{id}N_{id}} = K_{ir}(\mathbf{t}_{id}, \mathbf{t'}_{id})$; i.e., $\mathbf{g}_{ir} \sim \mathcal{GP}(\mathbf{0}, \mathbf{K}^{(ir)}_{N_{id}N_{id}})$ and $\mathbf{g}_{ir}\indep \mathbf{g}_{i'r'},\forall r\neq r',\forall i, i'$. The parameters of this kernel are shared across different signals. The signal-specific function, is generated from a GP whose kernel parameters are signal-specific: $\mathbf{v}_{id} \sim \mathcal{GP}(\mathbf{0}, \mathbf{K}^{(id)}_{N_{id}N_{id}})$.

We choose Mat\'ern-1/2 kernel for each latent function (see, e.g., \citet{Rasmussen2006}). For shared latent functions, for instance, we have
$K_{ir}(t, t') = \exp(-\frac{1}{2}\frac{|t-t'|}{l_{ir}})$,
where $l_{ir}>0$ is the length-scale of the kernel, and $|t-t'|$ is the Euclidean distance between $t$ and $t'$.

We assume $\epsilon_{id}(t)$ is generated from a non-standardized Student's t-distribution with scale $\sigma_{id}$ and 3 degrees of freedom, $\epsilon_{id}(t) \sim \mathcal{T}_3(0, \sigma_{id})$. We choose Student's t-distribution because it has heavier tail than Gaussian distribution and is more robust against outliers; see, e.g., \citet{jylanki2011robust}.

Intuitively, this particular structure of our model posits that the patterns exhibited by the multivariate time-series of each individual can be described by two components: a low-dimensional function space shared among all signals and a signal-specific latent function. The shared component is the primary mechanism for learning the correlations among signals; signals that are more highly correlated give high weights to the same set of latent functions (i.e., $w_{idr}$ and $w_{id'r}$ are similar).
Modeling correlations is natural in domains like health where deterioration in any single organ system is likely to affect multiple signals. Further, by modeling the correlations, the model can improve estimation when data are missing for a sparsely sampled signal based on the correlations with more frequently sampled signals.

 {In our experiments, we set $R=2$ and initialize the length-scales such that one kernel captures the short-term changes and the other learns the long-term trends in the shared latent functions. In general, kernel length-scales can be individual-specific free parameters. However, to capture common dynamic patterns and share statistical strength across individuals, we found it helpful to share the length-scale for each latent function across all individuals. One challenge in doing so is that individuals may have different length of observations and one length-scale may not fit all. Experimentally, we found that length-scale of the long-range kernel has a linear relation with the logarithm of the length of observation for each individual. To capture this relation, we define}
\begin{align}
l_{ir} = \mathscr{T}(\beta_r \log(\bar{t}_i) + \beta_{0r}), \forall r=1,2,...,R\, ,\label{lengthscale_eqn}
\end{align}
 {where $\bar{t}_i = \max_d\max_n t_{idn}$ is the maximum observed time for individual $i$, and $\beta_r$ and $\beta_{0r}$ are population-level parameters which we estimate along with other model parameters. Thus, instead of sharing the same length-scale between individuals who may have different length of observations, we share $\beta_r$ and $\beta_{0r}$. Also, $\mathscr{T}: \mathbb{R}\rightarrow \mathbb{R}^+$ is an appropriate mapping to obtain positive length-scale. We set $\mathscr{T}(x) = 0.1 + 15000/(1+\exp(-x))$ to obtain $l_{ir}\in[0.1, 15000]$; this prevents too small or too large length-scales.
We initialize $\beta_r=1$ and $\beta_{0r}=-12$ for the long-range kernel, yielding length-scales in the range of $131$-$635$ minutes for longitudinal data of duration $1440$-$7200$ (minutes). We also initialize $\beta_r=10^{-5}$ and $\beta_{0r}=-5$ for the short-term kernel to obtain initial length-scales of ${\sim}100$ minutes with minor dependence on the duration of the longitudinal data. After initialization, we learn these parameters along with other parameters of our model}\footnote{ {A more flexible formulation is possible by adding an individual-specific term: $l_{ir} = \mathscr{T}(\beta_r \log(\bar{t}_i) + \beta_{0r}+l'_{ir})$, with $l'_{ir}\sim \mathcal{N}(0, \tau^2)$ for some noise level $\tau^2$. In our experiments, we did not observe significant performance improvement by using this alternative formulation.}}.

We similarly define kernels and length-scales for signal-specific latent functions, $K_{id}(t, t') = \exp(-\frac{1}{2}\frac{|t-t'|}{l_{id}})$, with
$l_{id} = \mathscr{T}(\beta_d \log(\bar{t}_{id}) + \beta_{0d}), \forall d=1,2,...,D\,$,
where $\bar{t}_{id} = \max_n t_{idn}$, and $\beta_d$ and $\beta_{0d}$ are free parameters. We initialize $\beta_d=10^{-5}$ and $\beta_{0d}=-5$ to capture short-term signals-specific trends.

Unless there is ambiguity, we hereon drop the index for individual $i$.  Also, to simplify the notation, we assume $\mathbf t_{id} = \mathbf{t}_i, \forall d,$ and write $\bfK^{(r)}_{\bfN\bfN} = K_{ir}(\mathbf{t}_{i}, \mathbf{t}'_{i})$. We emphasize that the observations from different signals need not be aligned for our learning algorithm.

\subsection{Time-to-Event Sub-model}\label{t2e_submodel}
The time-to-event sub-model computes the event probabilities conditioned on the features $\bff^{0:t}$ which are estimated in the longitudinal sub-model. Specifically, given the predictions $\bff^{0:t}_i$ for each individual $i$ who has survived up to time $t$, we define a dynamic hazard function for time $s \geq t$:
\begin{align}
\lambda(s;t) = \exp(b + a(s-t) + \bfgamma^T \bfx_{t} + \bar{f}_{i}(t))\, , \forall s\geq t\, ,
\label{lambda_landmark}
\end{align}
where
\begin{align}
\bar{f}_i(t) &= \bfalpha^T \int_0^t \rho_c(t';t)\bff_i(t')\dee t'\, , \label{fbar_eqn}\\
\rho_c(t';t) &= c\frac{\exp(-c(t-t'))}{1-\exp(-ct)}\, , \forall t'\in [0,t]\, ,
\end{align}
and $\bff_i(t) \triangleq [f_{i1}(t), ..., f_{iD}(t)]^T$. Here, $\rho_c(t';t)$ is the weighting factor for the integral, and $c\geq 0$ is a free parameter. At any time $t$, $\rho_c(t';t)$ gives exponentially larger weight to most recent history of the feature trajectories; the parameter $c$ controls the rate of the exponential weight\footnote{ {We can also make $\rho_c$ signal-specific (with parameter $c_d$ for each $d$), to control the weight assigned to the history of each signal separately.}}. The relative weight given to most recent history increases by increasing $c$. We also normalize $\rho$ so that $\int_0^t \rho_c(t';t) \dee t' = 1, \forall t, c$.  {Similar ideas for incorporating signal histories have also been explored by \citet{rizopoulos2014combining}.}

We can also write the hazard function in terms of the latent functions by substituting (\ref{model}) into (\ref{lambda_landmark}):
\begin{align}
\lambda(s;t) &= \lambda_0(s;t)\exp\bigg(\bfgamma^T \bfx_{t} + \sum_{d=1}^D \kappa'_d \int_0^t \rho_c(t';t) v_d(t')\dee t' \nonumber \\
&~~~~~~+ \sum_{r=1}^R \omega'_r \int_0^t \rho_c(t';t) g_r(t')\dee t'\bigg)\, ,
\label{lambda_latent}
\end{align}
where $\kappa'_d \triangleq \kappa_d \alpha_d$, $\omega'_r \triangleq \sum_{d=1}^D \omega_{dr}\alpha_d$, and $\lambda_0(s;t)=\exp(b + a(s-t))$. In section \ref{inf_sec}, we describe how we can analytically compute the integrals of the latent functions in (\ref{lambda_latent}).
Given (\ref{lambda_latent}), at any point $t$, we compute the distribution of the event probability $p_H(h)$. For a given realization of $\bar{f}$, the event probability is:
\begin{align}
H(\Delta|\bar{f},t) = 1-\exp\big(-\lambda(t;t)\frac{1}{a}(e^{a\Delta}-1)\big)\, .
\label{landmark_failure_prob}
\end{align}

The hazard function defined in (\ref{lambda_landmark}) is based on linear features (i.e., $\exp(\bfalpha^T\int_0^t \rho_c(t';t)\bff_i(t')\dee t')$). Linear features are common in survival analysis because they are interpretable. In our application of interest, interpretable features are preferred over non-linear features that are challenging to interpret.
Non-linear features can be incorporated within our framework. Recently, there have been a number of useful proposals for doing so. For instance, \citet{Joensuu2012}, \citet{Saul2016a} and \citet{Fernandez2016} propose variants of GPs to learn more complex dependencies between the covariates and time-to-event data. \citet{Ranganath2016} use deep exponential families \cite{Ranganath2014a} to develop a latent representation of diverse, multivariate data  (e.g., continuous and count) with a Weibull link function to predict the time-to-event from the inferred latent representation. Though these papers focus on the cross-section setting (i.e., they do not tackle longitudinal data), their approach for learning representations of non-linear features can be incorporated as needed within the proposed framework.

\subsection{Learning and Inference}\label{inf_sec}
In this section, we describe learning and inference for the proposed joint model. Our model has global and local parameters. Global parameters, denoted by $\Theta_0$, are the parameters of the time-to-event model ($\bfalpha, \bfgamma, a, b, c$) and the parameters defining the kernel length-scales ($\beta_r, \beta_{0r}, \beta_d, \beta_{0d}$); i.e., $\Theta_0=\{\bfalpha, \bfgamma, a, b, c, \beta_r, \beta_{0r}, \beta_d, \beta_{0d}\}$. Our procedure is to update the local parameters for a minibatch of individuals independently, and use the resulting distributions to update the global parameter. Unlike classical stochastic variational inference procedures, our local updates are highly non-linear and we make use of gradient-based optimization inside the loop.

\subsubsection{Local parameters}
The key bottleneck for inference is the use of robust sparse GPs in the longitudinal sub-model. Specifically, due to matrix inversion, even in the univariate longitudinal setting, GP inference scales \emph{cubically} in the number of observations. To reduce this computational complexity, we develop our learning algorithm based on the sparse variational approach \cite{Titsias2009,Hensman2013,Hensman2015,Matthews2016}. Also, the assumption of heavy-tailed noise makes the model robust to outliers, but this means that the usual conjugate relationship in GPs is lost: the variational approach also allows approximation of the non-Gaussian posterior over the latent functions.

 {Specifically, we integrate out each Gaussian process latent function and posit a variational distribution to approximate its posterior.} The local parameters of our model, denoted by $\Theta_i$, comprise the variational parameters controlling these GP approximations, noise-scale, and inter-process weights $\omega, \kappa$. We make point-estimates of these parameters.

Our model involves multiple GPs: for each individual, there are $R$ latent functions $\mathbf g_r$ and $D$ signal-specific functions $\mathbf v_d$. In our variational approximation, each of these functions is assumed independent, and controlled by $M$ inducing input-response pairs $\bfZ, \bfu$, where $\bfZ$ are some pseudo-inputs (which we arrange on a regular grid) and $\bfu$ are the values of the process at these points  {with distribution $p(\bfu)$. We give the variables $\bfu_r$ a variational distribution $q(\bfu_r) = \mathcal{GP}(\bfm_r, \bfS_r)$ which gives rise to a variational GP distribution, $q(\mathbf g_r) = \int p(\mathbf g_r|\bfu_r)q(\bfu_r) \dee \bfu_r = \mathcal{GP}(\bfmu_{g_r}, \bfSigma_{g_r}), \forall r=1,...,R$}, where $\bfmu_{g_r} = \bfK^{(r)}_{\bfN \bfZ}\bfK^{{(r)}^{-1}}_{\bfZ\bfZ}\bfm_r$ and
\begin{align}
\bfSigma_{g_r} = \bfK^{(r)}_{\bfN\bfN}-\bfK^{(r)}_{\bfN\bfZ}\bfK^{{(r)}^{-1}}_{\bfZ\bfZ}(\mathbf I - \bfS_r\bfK^{{(r)}^{-1}}_{\bfZ\bfZ})\bfK^{(r)}_{\bfZ\bfN}\, , \nonumber
\end{align}
where $\bfK^{(r)}_{\bfN\bfZ} = K_r(\mathbf{t}, \bfZ)$. We similarly obtain the variational distribution  {$q(\mathbf v_d) = \int p(\mathbf v_d|\bfu_d)q(\bfu_d) \dee \bfu_d = \mathcal{GP}(\bfmu_{v_d}, \bfSigma_{v_d}), \forall d$}.

Since the functions of interest $\bff_d,\forall d=1,2,...,D,$ are given by linear combinations of these processes, the variational distribution $q(\bff)$ is given by taking linear combinations of these GPs. Specifically:
\begin{equation}
    q(\bff_d) = \mathcal{GP}(\bfmu_{d}, \bfSigma_d)\, ,
    \label{q_f}
\end{equation}
where $\bfmu_{d} = \sum_{r=1}^R \omega_{dr}\bfmu_{g_r} + \kappa_d\bfmu_{v_d}$ and $\bfSigma_{d} = \sum_{r=1}^R \omega^2_{dr}\bfSigma_{g_r} + \kappa^2_d\bfSigma_{v_d}$.
These variational distributions are crucial in computing  {the lower bound on marginal likelihood (evidence lower bound (ELBO))}, the objective function we use in optimizing the variational parameters $\bfm, \bfS$.

For each individual, we are given longitudinal data $\mathbf y_i$, time-to-event data $\mathbf T_i$, and censoring data $\delta_i$. Collecting these into $\mathcal D_i$, the likelihood function for an individual is $p(\mathcal D _i|\Theta_i, \Theta_0)$. Hereon,  {unless there is ambiguity}, we drop the individual subscript, $i$, and the explicit conditioning on $\Theta_i$ and $\Theta_0$.
Given the GP approximations and using Jensen's inequality, we obtain
\begin{align}
\log \int p(\mathcal{D}|\Theta, \bff)&p(\bff|\bfu)p(\bfu)\dee \bff \dee \bfu\label{p_D}\geq E_{q(\bff)} \big[\log p(\bfy|\bff)\\
&  + \log p(\mathbf{T}, \delta|\bff)\big]-\text{KL}(q(\bfu)||p(\bfu))=\text{ELBO}_i\, ,\nonumber
\end{align}
where $q(\bff) = E_{q(\bfu)} p(\bff|\bfu)$. In computing (\ref{p_D}), we used the fact that the time-to-event and longitudinal data are independent conditioned on $\bff$.

First consider computation of $E_{q(\bff)} \log p(\bfy|\bff)$.
Since conditioned on $\bff$, the distribution of $\bfy$ factorizes over $d$, we obtain $E_{q(\bff)} \log p(\bfy|\bff)=\sum_d E_{q(\bff_d)} \log p(\bfy_d|\bff_d)$, where $q(\bff_d)$ is computed in (\ref{q_f}). Given our choice of the noise distribution, we cannot compute this expectation analytically. However, conditioned on $\bff_d$, $\log p(\bfy_d|\bff_d)$ also factorizes over all individual observations. Thus, this expectation reduces to a sum of several one-dimensional integrals, one for each observation, which we easily approximate using Gauss-Hermite quadrature.

Next consider computation of $E_{q(\bff)}\log p(\mathbf{T}, \delta|\bff)$.
Unlike $\bfy$, likelihood of the time-to-event sub-model does not factorize over $d$. We also need to take expectations of the terms involving the hazard function (\ref{lambda_latent}) which requires computing integral of latent functions over time. To this end, we make use of the following property:
\begin{quote}
Let $f(t)$ be a Gaussian process with mean $\mu(t)$ and kernel function $K(t, t')$. Then, $\int_0^T \rho(t) f(t) \dee t$ is a Gaussian random variable with mean $\int_0^T \rho(t)\mu(t)\dee t$ and variance $\int_0^T \int_0^T \rho(t)K(t,t')\rho(t')\dee t \dee t'$.
\end{quote}
See Appendix \ref{appendA} for the proof of this property. Similar ideas have been used in Bayesian quadrature \cite{Osborne2012}, inter-domain sparse GPs \cite{Lazaro-Gredilla2009}, and Fourier features for sparse GPs \cite{Hensman2016}.

Using this property, we can easily show that $\bar{f}_i(t)=\bfalpha^T\int_0^t\rho_c(t';t)\bff_i (t')\dee t'$ is a Gaussian random variable with mean $\mu^{(t)}_i$ and variance $\sigma_i^{2^{(t)}}$, which we compute analytically in closed form. We then compute $E_{q(\bff)}\log p(\mathbf{T}, \delta|\bff)$ by replacing the likelihood function as defined in (\ref{t2e_lkh}) and following the dynamic approach for defining the hazard function described in section \ref{t2e_submodel}. Expectation of the term related to interval censoring in the likelihood function is not available in closed form. Instead, we compute Monte Carlo estimate of this term and use reparameterization tricks \cite{Kingma2013} for computing gradients of this term with respect to model parameters. Detailed derivations are given in Appendix \ref{appendA}.

Now, we can compute $\text{ELBO}_i$ in (\ref{p_D}). The KL term in (\ref{p_D}) is available in closed form.

\subsubsection{Global parameters}
Here, we describe estimation of the global parameters $\Theta_0=\{\bfalpha, \bfgamma, a, b, c, \beta_r, \beta_{0r}, \beta_d, \beta_{0d}\}$. The overall objective function for maximizing $\Theta_0$ is: $\text{ELBO}=\sum_{i}^I\text{ELBO}_i$ where $I$ is the total number of individuals. Since ELBO is additive over $I$ terms, we can use stochastic gradient techniques. At each iteration of the algorithm, we randomly choose a mini-batch of individuals and optimize $\text{ELBO}$ with respect to their local parameters (as discussed in section 3.3.1), keeping $\Theta_0$ fixed.
We then perform one step of stochastic gradient ascent based on the gradients computed on the mini-batch to update global parameters. We repeat this process until either relative change in global parameters is less than a threshold or maximum number of iterations is reached. We use AdaGrad \cite{Byrd1995} for stochastic gradient optimization.

\subsubsection{Computational complexity:}
 {Computing the variational GP approximation for each latent function requires inverting an $M\times M$ matrix, which scales cubically in $M$ (i.e. the number of inducing points), and multiplying $N\times M$ and $M \times M$ matrices, with complexity $\mathcal{O}(NM^2)$. Therefore, the overall complexity of inference is $\mathcal{O}\big((R+D)(M^3+NM^2)\big)$. Typically, $M \ll N$ which yields an overall complexity of $\mathcal{O}\big((R+D)NM^2\big)$.}

\section{Uncertainty-Aware Event Prediction}\label{uncertainty_section}\label{robust_section}
The joint model developed in section \ref{methods_section} computes the probability of occurrence of the event $H(\Delta|\bar{f},t)$ within any given horizon $\Delta$. Here, we derive the optimal policy that uses this event probability and its associated uncertainty to detect occurrence of the event. The desired behavior for the detector is to wait to see more data and abstain from classifying when the estimated event probability is unreliable and the risk of incorrect classification is high. To obtain this policy, we take a decision theoretic approach \cite{poor2013introduction}.

At any given time, the detector takes one of the three possible actions: it makes a positive prediction (i.e., to predict that the event will occur within the next $\Delta$ hours), negative prediction (i.e., to determine that the event will not occur during the next $\Delta$ hours), or abstains (i.e., to not make any prediction). The detector decides between these actions by trading off the cost of incorrect classification against the penalty of abstention. We define a risk (cost) function by specifying a relative cost term associated with each type of possible error (false positive and false negative) or abstention. We then derive an optimal decision function (policy) by minimizing the specified risk function.

Specifically, for every individual $i$, given the observations up to time $t$, our goal is to determine whether the event will occur ($\psi_i=1$) within the next $\Delta$ hours or not ($\psi_i=0$). Hereon, we again drop the $i$ and $t$ subscripts for brevity. We treat $\psi$ as an unobserved Bernoulli random variable with probability $Pr(\psi=1)=H(\Delta|\bar{f},t)$. Our joint model estimates this probability by computing the distribution $p_H(h)$. Detailed derivations of this distribution are provided in Appendix \ref{appendB}. The distribution on $H$ provides valuable information about the uncertainty around the estimate of $Pr(\psi=1)$. Our robust policy, which we derive next, uses this information to improve reliability of event predictions.

We denote the decision made by the detector by $\hat{\psi}$. The optimal policy chooses an action $\hat{\psi}\in\{0, 1, a\}$, where $a$ indicates abstention, and $\hat{\psi}=0,1$, respectively, denote negative and positive prediction.

We specify the risk function by defining $L_{01}$ and $L_{10}$, respectively, as the cost terms associated with false positive (if $\psi=0$ and $\hat{\psi}=1$) and false negative (if $\psi=1$ and $\hat{\psi}=0$) errors and defining $L_a$ as the cost of abstention (if $\hat{\psi}=a$). Conditioned on $\psi$, the overall risk function is
\begin{align}
R(\hat{\psi};\psi) &= \mathds{1}(\hat{\psi}=0) \psi L_{10} + \mathds{1}(\hat{\psi}=1) (1-\psi) L_{01} \nonumber\\
&+ \mathds{1}(\hat{\psi}=a) L_a\, ,
\label{risk_given_delta}
\end{align}
where the indicator function, $\mathds{1}(x)$, equals 1 or 0 according to whether the boolean variable $x$ is true or false.

Since $\psi$ is an unobserved random variable, instead of minimizing (\ref{risk_given_delta}), we should minimize the expected value of $R(\hat{\psi};\psi)$ with respect to the distribution of $\psi$, $Pr(\psi=1)=H$: i.e., $R(\hat{\psi};H) = \mathds{1}(\hat{\psi}=0) H L_{10} + \mathds{1}(\hat{\psi}=1) (1-H) L_{01} + \mathds{1}(\hat{\psi}=a) L_a$.
Because $H$ is a random variable, the expected risk function $R(\hat{\psi};H)$ is also a random variable for every possible choice of $\hat{\psi}$. The distribution of $R(\hat{\psi};H)$ can be easily computed based on the distribution of $H$, $p_H(h)$.

We obtain the robust policy by minimizing the \textit{quantiles} of the risk distribution. Intuitively, by doing this, we minimize the maximum cost that could occur with a certain probability. For example, with probability $0.95$, the cost under any choice of $\hat{\psi}$  is less than $R^{(0.95)}$, the 95th quantile of the risk distribution  $R(\hat{\psi};H)$.

Specifically, let $h^{(q)}$ be the q-quantile of the distribution $p_H(h)$; i.e., $\int_{0}^{h^{(q)}}p_H(h)\dee h = q$. We compute the q-quantile of the risk function, $R^{(q)}(\hat{\psi})$, for $\hat{\psi}=0, 1$, or $a$:

When $\hat{\psi}=0$, the q-quantile of the risk function is $L_{10}h^{(q)}$. Similarly, for the case of $\hat{\psi}=1$, the q-quantile of the risk function is $L_{01}h^{(1-q)}$. Here, we use the property that the q-quantile of the random variable $1-H$ is $1-h^{(1-q)}$, where $h^{(1-q)}$ is the (1-q)-quantile of $H$ (see Appendix \ref{appendB} for details). Finally, q-quantile of the risk function is $L_a$ in the case of abstention ($\hat{\psi}=a$).
We obtain the q-quantile of the risk function:
\begin{align}
R^{(q)}(\hat{\psi}) &= \mathds{1}(\hat{\psi}=0) h^{(q)} L_{10} + \mathds{1}(\hat{\psi}=1) (1-h^{(1-q)}) L_{01} \nonumber \\
&+ \mathds{1}(\hat{\psi}=a) L_a\, .
\label{risk_quantile}
\end{align}

We minimize (\ref{risk_quantile}) to compute the optimal policy. The optimal policy determines when to choose $\hat{\psi}=0,1$, or $a$ as a function of $h^{(q)}$, $h^{(1-q)}$, and the cost terms $L_{01}$, $L_{10}$, and $L_a$. In particular, we should choose $\hat{\psi}=0$ when $h^{(q)} L_{10} \leq (1-h^{(1-q)}) L_{01}$ and $h^{(q)} L_{10} \leq L_a$.
Because the optimal policy only depends on the relative cost terms, to simplify the notation, we define $L_1 \triangleq \frac{L_{01}}{L_{10}}$ and $L_2 \triangleq \frac{L_a}{L_{10}}$. Further, we assume that $q>0.5$ and define $c_q \triangleq h^{(q)}-h^{(1-q)}$. Here, $c_q$ is the $1-2q$ confidence interval of $H$. Therefore, substituting $L_1, L_2$, and $c_q$, the condition for choosing $\hat{\psi}=0$ simplifies to $h^{(q)}\leq L_1(1+c_q)/(1+L_1)$ and $h^{(q)}\leq L_2$.

We similarly obtain optimal conditions for choosing  $\hat{\psi}=1$ or $\hat{\psi}=a$. The optimal decision rule is given as follows:
\begin{align}
\hat{\psi} =
\begin{cases}
0,& \text{if } h^{(q)}\leq \ubar{\tau}{(c_q)},\\
1,& \text{if } h^{(q)}\geq \obar{\tau}(c_q),\\
a,& \text{if } \ubar{\tau}(c_q) < h^{(q)} < \obar{\tau}(c_q),\\
\end{cases}
\label{delta_r}
\end{align}
where $\ubar{\tau}(c_q) = \min\{L_1\frac{1+c_q}{1+L_1}, L_2\}$ and $\obar{\tau}(c_q) = \max\{L_1\frac{1+c_q}{1+L_1},1+c_q-\frac{L_2}{L_1}\}$.

\begin{figure}[t]
\centering
\includegraphics[scale=1]{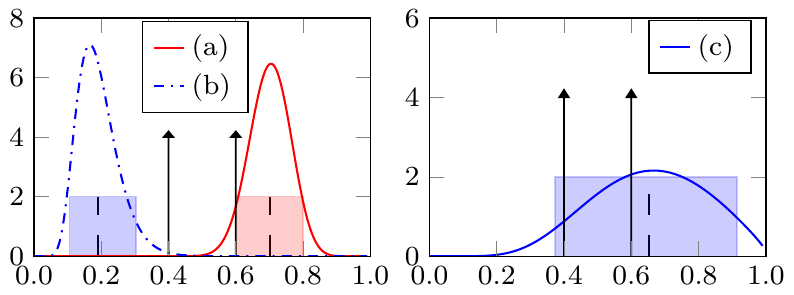}
\caption{Three example decisions made using the policy described in Fig. \ref{policy_alg} with $L_1=1$ and $L_2=0.4$. The shaded area is the confidence interval $[h^{(1-q)}, h^{(q)}]$ for some choice of $q$ for the three distributions, (a), (b), and (c). The arrows at $0.4$ and $0.6$ are $L_2$ and $1-\frac{L_2}{L_1}$, respectively. All cases satisfy $c_q\geq L_2\frac{1+L_1}{L_1}-1$. The optimal decisions are $\hat{\psi}=1$ for (a), $\hat{\psi}=0$ for (b), and $\hat{\psi}=a$ for (c).} \label{policy_fig}
\end{figure}

The thresholds $\ubar{\tau}(c_q)$ and $\obar{\tau}(c_q)$ in (\ref{delta_r}) can take two possible values depending on how $c_q$ is compared to $L_1$ and $L_2$:
in the special case that $c_q > L_2\frac{1+L_1}{L_1}-1$, the prediction is made by comparing the confidence interval $[h^{(1-q)}, h^{(q)}]$ against thresholds $L_2$ and $1-\frac{L_2}{L_1}$. In particular, if the entire confidence interval is above $1-\frac{L_2}{L_1}$ (i.e., if $h^{(1-q)}>1-\frac{L_2}{L_1}$ as shown in Fig. \ref{policy_fig}a), we predict $\hat{\psi}=1$. If the entire confidence interval is below $L_2$ (i.e., if $h^{(q)}<L_2$ as shown in Fig. \ref{policy_fig}b), we declare $\hat{\psi}=0$.  And if none of these conditions are met, the classifier abstains from making any decision (as shown in Fig. \ref{policy_fig}c).
In the case of $c_q < L_2\frac{1+L_1}{L_1}-1$ (i.e., the uncertainty level is below a threshold), $\hat{\psi}$ is 0 or 1, respectively, if $h^{(q)}+L_1h^{(1-q)}$ is less than or greater than $L_1$.
We summarize this policy in Fig. \ref{policy_alg}.

In principle, the cost terms $L_1$, $L_2$, and $q$ are provided by the field experts based on their preferences for penalizing different types of error and their desired confidence level. Alternatively, one could perform a grid search on $L_1, L_2, q$ and choose the combination that achieves the desired performance with regard to specificity, sensitivity and the false alarm rates. In our experiments, we take the latter approach.
\begin{algorithm}[t]
\renewcommand{\thealgorithm}{}
\captionof{figure}{Robust Prediction Policy}\label{policy_alg}
\begin{algorithmic}[1]
\STATE Input: $1-2q$ confidence interval ($[h^{(1-q)}, h^{(q)}]$) of the event probability $H$. Let $c_q = h^{(q)}-h^{(1-q)}$, $q > 0.5$. Also, $L_1 \triangleq \frac{L_{01}}{L_{10}}$ and $L_2 \triangleq \frac{L_a}{L_{10}}$, where $L_{01},L_{10},$ and $L_a$ are the cost of false positive, false negative, and abstention, respectively.
\STATE Output: $\hat{\psi}\in\{0, 1, a\}$.
\STATE If $ c_q \geq L_2\frac{1+L_1}{L_1}-1$: (large confidence interval - high uncertainty)\\
\hspace{5mm}Set $\hat{\psi}=0$ if $h^{(q)}\leq L_2$.\\
\hspace{5mm}Set $\hat{\psi}=1$ if $h^{(1-q)} \geq 1-\frac{L_2}{L_1}$ .\\
\hspace{5mm}Set $\hat{\psi}=a$ otherwise.
\STATE If $ c_q < L_2\frac{1+L_1}{L_1}-1$: (small confidence interval - low uncertainty)\\
\hspace{5mm}Set $\hat{\psi}=0$ if $h^{(q)} + L_1 h^{(1-q)} < L_1$.\\
\hspace{5mm}Set $\hat{\psi}=1$ if $h^{(q)} + L_1 h^{(1-q)} \geq L_1$ .\\
\end{algorithmic}
\end{algorithm}

\subsection{Special Case: Policy without Uncertainty Information}
\label{special-case:point-estimate}
Imputation-based methods and other approaches that do not account for the uncertainty due to missingness can only compute point-estimates of the failure probability, $H$. In that case, we can think of the distribution over $H$ as a degenerate distribution with mass 1 on the point estimate of $H$; i.e., $p_H(h) = \mathds{1}(h-h_0)$, where $h_0$ is the point estimate of $H$. Here, because of the degenerate distribution, we have $h^{(q)} = h^{(1-q)} = h_0$ and $c_q=0$.

In this special case, the robust policy summarized in Fig. \ref{policy_alg} reduces to the following simple case:
\begin{align}
\hat{\psi} =
\begin{cases}
0,& \text{if } h_0\leq \ubar{\tau},\\
1,& \text{if } h_0\geq \obar{\tau},\\
a,& \text{if } \ubar{\tau} < h_0 < \obar{\tau},\\
\end{cases}
\label{delta_v}
\end{align}
where $\ubar{\tau} = \min\{L_2, \frac{L_1}{1+L_1}\}$ and $\obar{\tau} = \max\{1-\frac{L_2}{L_1}, \frac{L_1}{1+L_1}\}$. This policy is similar to classification with abstention framework introduced in \citet{chow1957optimum}.

As an example, consider the case that $L_1=1$. Here, if the relative cost of abstention is $L_2\geq 0.5$, we have $\obar{\tau}=\ubar{\tau} = 0.5$, which is the policy for a binary classification with no abstention and a threshold equal to $0.5$. Alternatively, when $L_2 < 0.5$, the abstention interval is $[L_2, 1-L_2]$. In this case, the classifier chooses to abstain when the event probability $L_2 < h_0 < 1-L_2$ (i.e., when $h_0$ is close to the boundary).

\subsubsection{Comparison with the robust policy with uncertainty:}
Both the robust policy (\ref{delta_r}) and its special case (\ref{delta_v}) are based on comparing a statistic with an interval, i.e., $h^{(q)}$ with the interval $[\ubar{\tau}(c_q), \obar{\tau}(c_q)]$ in the case of (\ref{delta_r}), and $h_0$ with the interval $[\ubar{\tau}, \obar{\tau}]$ in the case of (\ref{delta_v}).

An important distinction between these two cases is that, under the policy (\ref{delta_v}), the abstention region only depends on $L_1$ and $L_2$ which are the same for all individuals, but under the robust policy (\ref{delta_r}), the length of the abstention region is $\max\{0, 1 + c_q -L_2\frac{1+L_1}{L_1}\}$. That is, the abstention region adapts to each individual based on the length of the confidence interval for the estimate of $H$. The abstention interval is larger in cases where the classifier is uncertain about the estimate of $H$. This helps to prevent incorrect predictions. For instance, consider example (c) in Fig. \ref{policy_fig}. Here the expected value $h_0$ (dashed line) is greater than $\obar{\tau}$ but its confidence interval (shaded box) is relatively large. Suppose this is a negative sample, making a decision based on $h_0$ (policy (\ref{delta_v})) will result in a false positive error. In order to abstain on this individual under the policy (\ref{delta_v}), the abstention interval should be very large. But because the abstention interval is the same for all individuals, making the interval too large leads to abstaining on many other individuals on whom the classifier may be correct.
Under the robust policy, however, the abstention interval is adjusted for each individual based on the confidence interval of $H$. In this particular case, for instance, the resulting abstention interval is large (because of large $c_q$), and therefore, the false positive prediction is avoided.

\vspace{-2mm}
\section{Related Work}\label{related_work}
\textbf{Joint models for longitudinal and event data:} Our proposed model builds upon the extensive prior literature on \emph{joint models} for longitudinal and time-to-event data. Here, a joint probability distribution is posited on the longitudinal and time-to-event data. For example, \citet{Rizopoulos2011a,Rizopoulos2010} uses generalized mixed-effects models for modeling the longitudinal data and computes the time-to-event distribution conditioned on the mean predictions from the longitudinal model. \citet{Proust-Lima2014} propose a more flexible joint model where an individual's data are assumed to be generated from one of a fixed number of classes and the longitudinal data from any individual class are modeled using a polynomial function. Coefficients from the longitudinal model act as predictors for the time-to-event distribution. While these models---by jointly modeling the longitudinal and event data---provide a principled way for propagating uncertainty due to missingness in estimating event probabilities, \textit{their applicability to challenging new domains {such as clinical data} is limited by the need to make strong parametric assumptions} about the form of the longitudinal data.

More recently, others have introduced more flexible ways to represent the longitudinal data. For example, \citet{Proust-Lima2016} extends their work discussed above in latent class modeling to include more flexible forms for the longitudinal data: specifically, for a given class, multiple longitudinal signals are correlated through a shared latent process which is  modeled as a Gaussian process with the mean represented by a linear mixed-effects model. Inference for this model scales \textit{cubically} in the number of unique time-points where observations are obtained, i.e., $\mathcal{O}(N^3)$. \citet{futoma2016} leverage flexible semi-parametric models introduced by \citet{Schulam2015} for modeling canonical progression patterns in the longitudinal data. Their approach also scales cubically in the number of observations $\mathcal{O}(N^3)$. Further, their work focuses on the setting with a single longitudinal marker and assumes alignment across time series from multiple individuals.

\textbf{Two-Stage Approaches:} Instead of jointly modeling the longitudinal and time-to-event data, one can take a two-stage approach.
Here, the most common approach is to use \textit{imputation} to fill in the missing data \cite{Enders2010,little2014statistical} and then apply time-to-event techniques on the completed data; e.g., \cite{Henry2015}. Some commonly used imputation techniques for longitudinal data are mean substitution, last-observation-carried-forward, and regression imputation \cite{Enders2010,little2014statistical}. In the latter approach, for instance, a regression model is used to impute values for a missing feature given other observed covariates. More sophisticated methods for imputation (e.g., by modeling the time series) can also be used.

A major drawback of these imputation methods that fill in missing data with a single substituted value is that they cannot propagate the error in imputing the missing values towards estimating the event probabilities. Multiple-imputation (MI) techniques circumvent this shortcoming \cite{Enders2010,little2014statistical}: they impute multiple values for each missing data point by sampling from the posterior distribution of the missing point given the observed data. This creates multiple completed datasets and the posterior uncertainty is quantified by averaging across these datasets. MI methods, however, suffer from the curse of dimensionality when applied to high-dimensional multivariate longitudinal signals with many irregularly sampled observations \cite{allison2001missing,Young2015}.

\textbf{Classification from Irregular Time Series:} Alternatively, one can treat event forecasting as a time-series classification task. Since this literature is vast, we briefly review relevant irregular time-series classification approaches focusing on clinical data.

Here, typically imputation methods or probabilistic models are used for extracting point estimates of features \cite{Ghassemi2015,stanculescu2014autoregressive,Liu2016}. For example, {\citet{Ghassemi2015} use multi-task Gaussian processes to model multiple longitudinal series and use features estimated from the resulting fitted data to predict occurrence of an event. The multi-task GP model used by \citet{Ghassemi2015} is also known as the intrinsic correlation model, which assumes that within-signal correlation structure is the same for all signals. }
Similarly, \citet{Alaa2016} use multi-task GPs for computing risk scores for patients in intensive care units. These approaches, however, do not have a principled mechanism to incorporate the uncertainty due to the missing longitudinal data in event prediction. Further, their method does not scale well to multivariate signals with many observations. Specifically, the computational cost of fitting their model grows \textit{cubically} in the number of signals ($D$) and the number of observations per signal ($N$), i.e., $\mathcal{O}(N^3D^3)$. This cost is prohibitive when either $N$ or $D$ is large.

\citet{Lasko2013} use GPs to model \textit{univariate} longitudinal data and train autoencoders on the GP predictions to extract more expressive nonlinear features for classification.
Parametric approaches such as hidden Markov models and linear dynamical systems have also been used for feature computation from clinical time-series for downstream time-series classification tasks (e.g., \cite{quinn2009factorial,Wuocw138,Liu2016}). Non-probabilistic methods based on recurrent neural networks have also been used for modeling irregularly sampled time series \cite{Lipton2016}. Again, these method generally lack a proper mechanism for incorporating uncertainty associated with the missing data. Other approaches exist for modeling event streams---e.g., piecewise-constant conditional intensity models (PCIMs) \cite{gunawardana2011model} model dependency in the timing of events across multiple discrete event types but they do not model continuous-valued time series.

\textbf{Reliable Prediction and Classification:} Accounting for uncertainty in \textit{training} a classifier has been investigated before. For instance, \citet{Li2016} proposed a framework for classification of (univariate) irregularly sampled time series using GPs. They use the estimates from a GP evaluated at a set of grid points as the features in a classifier. To account for uncertainty due to missingness, during training, they optimize the expected loss. However, during prediction, they do not incorporate it in individual classification decisions. In contrast, rather than only optimizing the expected loss, by taking into account the quantiles of the distribution, \emph{our policy leverages the shape of the event occurrence distribution at test time}. More specifically, using the uncertainty associated with the event probability, the proposed policy chooses when to wait and collect more samples before making a decision.

Classification with abstention has also been investigated before (see, e.g.,  \cite{chow1957optimum,golfarelli1997error,bartlett2008classification}). Deciding between abstention or classification in these methods is based on point-estimates of the event probabilities (i.e., these approaches provide policies akin to the policy described in Section \ref{special-case:point-estimate}). Unlike these methods, our approach incorporates the uncertainty in event probabilities in the form of confidence intervals.

\citet{Parrish2013} proposed a framework for reliable classification with incomplete data. Their notion of reliability is different from ours: they focus on the setting where each sample (e.g., video) belongs to a single class; reliable classification entails predicting the class of the sample from a partial sequence of frames such that the decision remains stable after observing the complete sample. \citet{Sangnier2016} and Hoai et al. \cite{Hoai2014} similarly exploit this monotonicity property in training classifiers for video classification. These works are different from ours in two key ways. First, their definition of reliability only holds when the time series are segmented into episodes containing a single event. Second, they do not consider settings with missing data.

\vspace{-2mm}
\section{Experimental Results}\label{exp_section}

We evaluate the proposed framework on the task of predicting \emph{when patients in the hospital are at high risk for septic shock}---a life-threatening adverse event. Currently, clinicians have only rudimentary tools for real-time, automated prediction for the risk of shock (see review of past work by \citet{Henry2015}). These tools suffer from high false alert rates. Early identification gives clinicians an opportunity to investigate and provide timely remedial treatments \cite{kumar2006duration}.

\subsection{Data}
We use the MIMIC-II Clinical Database \cite{GoldbergerAL2000}, a publicly available database, consisting of clinical data collected from patients admitted to a hospital (the Beth Israel Deaconess Medical Center in Boston). To annotate the data, we used the definitions described by \citet{Henry2015} for septic shock.
Censoring is a common issue in this dataset: patients for high-risk of septic shock can receive treatments that delay or prevent septic shock. In these cases, their true event time (i.e. event under no treatment) is \emph{censored} or unobserved. Following the approach of \cite{Henry2015}, we treat patients who received treatment and then developed septic shock as interval-censored because the exact time of shock onset could be at any time between the time of treatment and the observed shock onset time. Patients who never developed septic shock after receiving treatment are treated as right-censored. For these patients, the exact shock onset time could have been at any point after the treatment.

We model the following 10 longitudinal streams, which are the key clinical signals found to be highly predictive of septic shock by \citet{Henry2015}:
heart rate (HR), systolic blood pressure (SBP), urine output per Kg, respiratory rate (RR), Blood Urea Nitrogen (BUN), creatinine (CR), Glasgow coma score (GCS), blood pH as measured by an arterial line (Arterial pH), partial pressure of arterial oxygen (PaO2), and white blood cell count (WBC).
In addition, based on \cite{Henry2015}, we also include the following time-varying and time-invariant observed features that were found to be significant for identifying septic shock: time since first antibiotics, time since organ failure, and status of chronic liver disease, chronic heart failure, and diabetes.

 {We sub-sampled the original MIMIC-II database to include patients with at least 2 measurements per signal. This is not a technical requirement of the proposed model. Many of the baseline methods, described next, cannot naturally handle signals with very few or no measurements, and as a result perform poorly. This inclusion criterion is chosen to allow comparing against the baselines at their reasonable operating point. We then sub-sampled the patients with no septic shock to maintain the same ratio of septic shock as in the original cohort at 12-14\%. This yields a dataset of 3151 patients\footnote{The original cohort used by \citet{Henry2015} has 16,234 patients with 2291 septic shock patients.}. We randomly divided the patients into train (75\%) and test (25\%) sets, ensuring the same ratio of septic shock (${\sim}12\%$) in both. The training set consists of $2363$ patients, including $287$ patients with observed septic shock and $2076$ event-free patients. Further, of the patients in the training set, $279$ received treatment for sepsis, $166$ of which later developed septic shock (therefore, they are interval censored); the remaining $113$ patients are right censored. The test set consists of $788$ patients, $101$ with observed shock and $687$ event-free patients.}

 {For each test patient, we make predictions at 5 evaluation points. These are spaced equally over the two-day interval ending $15$ minutes prior to the time of shock onset, censoring, or the end of their hospital stay. We choose this setting because monitoring and early warning applications (the task considered in this paper) require frequent evaluations of the patient risk at multiple time points leading up to the event. This is different from standard time series classification tasks where the prediction is made once given the entire time series data. However, for the purpose of evaluation, in this paper, to avoid reporting bias from patients with very long hospital stays we choose to make predictions at 5 points for every patient. }

\begin{figure*}
\centering
\begin{subfigure}{0.84\textwidth}
\centering
\includegraphics[scale=1]{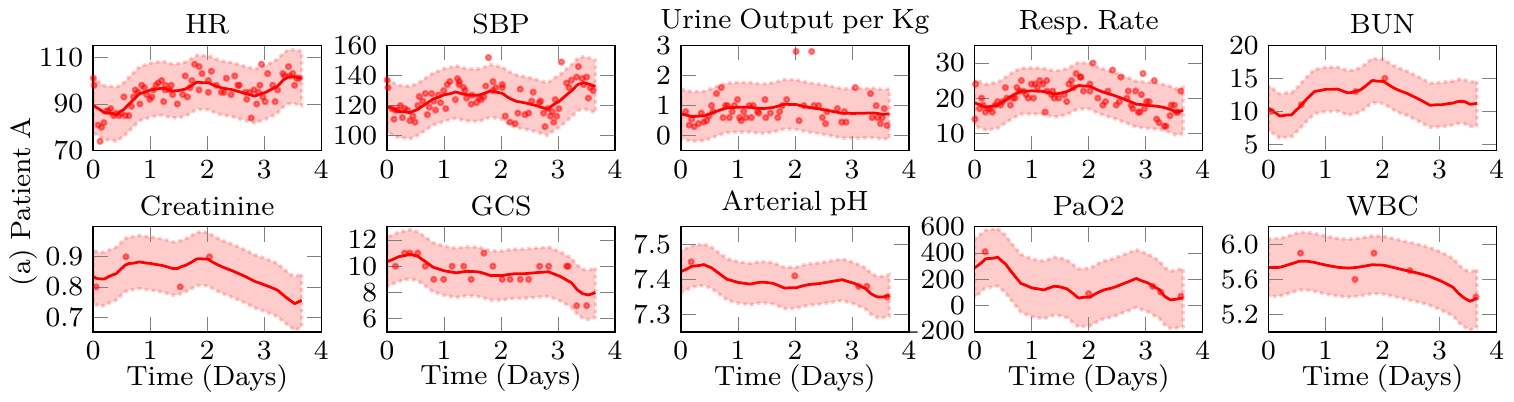}\vspace{-1mm}
\end{subfigure}\hfill
\begin{subfigure}{0.16\textwidth}
\centering
\includegraphics[scale=1]{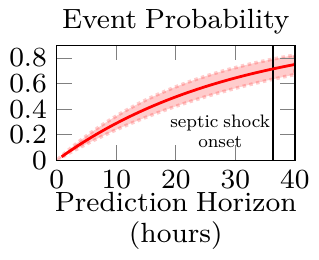}\vspace{-1mm}
\end{subfigure}\vspace{-0.0mm}

\begin{subfigure}{0.84\textwidth}
\centering
\includegraphics[scale=1]{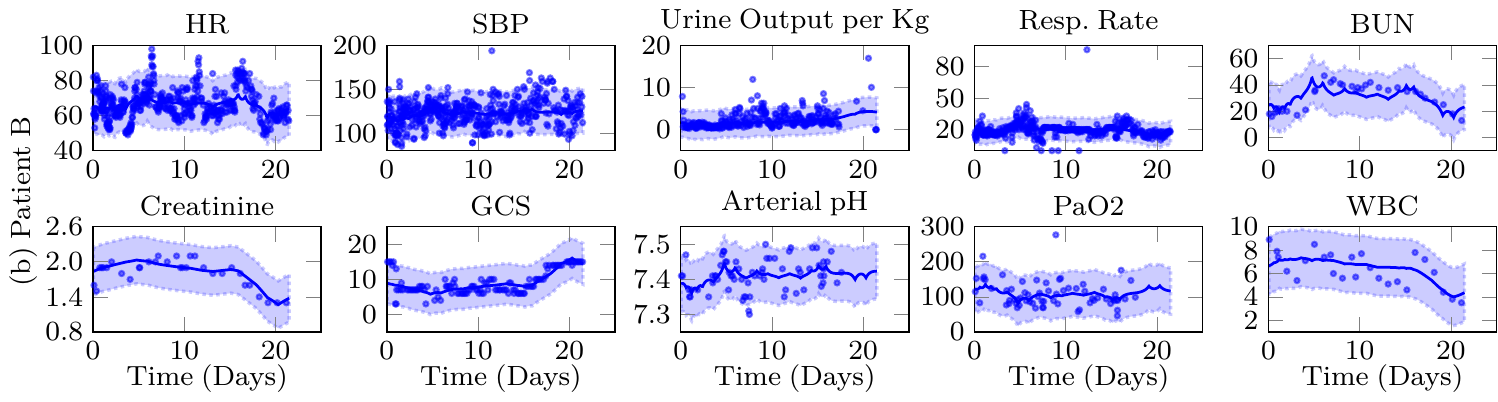}\vspace{-2mm}
\end{subfigure}\hfill
\begin{subfigure}{0.16\textwidth}
\centering
\includegraphics[scale=1]{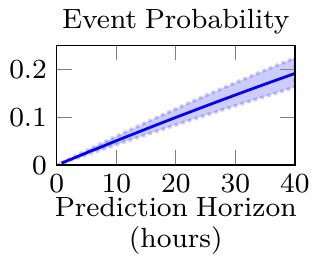}\vspace{-2mm}
\end{subfigure}
\caption{Data from 10 signals (dots) and longitudinal fit (solid line) along with their confidence intervals (shaded area) for two patients, (a) patient A with septic shock and (b) patient B with no observed shock. On the right, we show the estimated event probability for the following 40 hour period conditioned on the longitudinal data for each patient shown on the left.  {Septic shock for patient A occurs on day 5 of the stay. J-LTM observes the first 3.5 days of the longitudinal data from this patient and predicts the shock 36 hours before its onset.}}\label{sample_long}
\end{figure*}

We emphasize two challenging aspects of this data. First, individual patients have as many as {$2500$ observations per signal}. This is several \textit{orders of magnitude larger than the size of data that existing state-of-the-art joint models can handle} (past works tackled datasets containing $1$-$3$ signals with $10$-$50$ measurements each \cite{futoma2016,Proust-Lima2016}). Second, as shown in Fig. \ref{sample_long}, these signals have challenging properties: non-Gaussian noise, some are sampled more frequently than others, the sampling rate varies widely even within a given signal, and individual signals contain structure at multiple scales.

\subsection{Baselines}
To understand the benefits of the proposed model, we compare with the following commonly used alternatives.

\textit{1) MoGP:} For the first baseline, we implement a two-stage  {joint modeling} approach for modeling the longitudinal and time-to-event data. Specifically, we fit a  {multi-output GP (MoGP)} which provides highly flexible fits for imputing the missing data. \citet{Ghassemi2015} have shown state-of-the-art performance for modeling physiologic data using multivariate GP-based models. But, as previously discussed (see sections \ref{related_work} and \ref{methods_section}), their inference scales \textit{cubically} in the number of recordings; thus, making it impossible to fit to a dataset of our size. Here, we use the GP approximations described in section \ref{methods_section} for learning and inference. We use the mean predictions from the fitted MoGP to compute features for the hazard function (\ref{lambda_landmark}).  {The time-to-event model used for this baseline is similar to the model used for the proposed approach. The key difference is that due to the two-stage training approach, MoGP cannot propagate the uncertainty in the latent functions to the time-to-event component.} Using this baseline, we assess the extent to which a robust policy---that accounts for uncertainty due to the missing longitudinal data in estimating event probabilities---contributes to improving prediction performance.

\textit{2) JM:}  {For the second baseline, we use a two-stage joint model with a random-effects regression model for the longitudinal data. We fit a B-spline regression model with 20 knots independently to each signal of every patient to complete the missing data, and used the imputed values to compute the features for the hazard function. We also placed a population level Gaussian prior with diagonal covariance on the regression coefficients. The time-to-event component is similar to the one used for the proposed approach.}

\textit{3) Logistic Regression:} For this baseline, we use a time-series classification approach. Recordings from each time series signal are binned into 4-hour windows; for bins with multiple measurements, we use the average value. For bins with missing values, we use covariate-dependent (age and weight) regression imputation. Binned values from $10$ consecutive windows  {(i.e. the 40 hours preceding the time of prediction)} for all signals are used as features in a logistic regression (LR) classifier for event prediction. L2 regularization is used for learning the LR model; the regularization weight is selected using $2$-fold cross-validation on the training data.

\textit{4) SVM:} As another baseline, we replace the LR with an SVM to experiment with a more flexible classifier. We use the RBF kernel and determine hyperparameters using $2$-fold cross-validation on the training data.

\textit{5) RNN:}  {For this baseline, we train a recurrent neural network (RNN) on the binned multivariate time series using the prior ten 4-hour windows to predict the outcome variable (whether or not a patient will have septic shock).}

All of the baseline methods provide a point-estimate of the event probability at any given time. Thus, they use the special case of the robust policy with no uncertainty (policy (\ref{delta_v})) for event prediction.

\textbf{Evaluation:}
 {For all patients in the test set, we make predictions at each of the given evaluation points. For evaluation, we treat each prediction independently and aggregate the predictions across all evaluation points for all patients. From these, we compute the true positive rate (TPR), false positive rate (FPR), and the positive predictive value (precision) (PPV) as follows:}\vspace{-1mm}
\begin{align}
&\text{TPR} = \frac{\sum_{i}\mathds{1}(\hat{\psi}_i= 1,\psi_i=1)}{\sum_{i}\mathds{1}(\psi_i=1)}, \text{FPR} = \frac{\sum_{i}\mathds{1}(\hat{\psi}_i= 1,\psi_i=0)}{\sum_{i}\mathds{1}(\psi_i=0)}\,, \nonumber\\
&\text{PPV} = \big({\sum_{i}\mathds{1}(\hat{\psi}_i= 1,\psi_i=1)}\big)/\big({\sum_{i}\mathds{1}(\hat{\psi}_i=1)}\big)\, ,\vspace{-5mm}
\label{tpr_eqn}
\end{align}
We also compute the decision rate as the number of instances on which the classifier chooses to make a decision; i.e., $\big(\sum_{i}\mathds{1}(\hat{\psi}_i \neq a)\big)/\big(\sum_{i}1\big)$.  {Note that every classifier may abstain on a different set of prediction points. To make a fair comparison between the different methods, as shown in Eq. (\ref{tpr_eqn}), we compute the TPR and FPR rates with respect to all prediction points over all patients rather than the subset of points on which each classifiers chooses to make predictions; specifically, we compute the TPR with respect to $\sum_{i}\mathds{1}(\psi_i=1)$ rather than $\sum_{i}\mathds{1}(\psi_i=1, \hat{\psi}\neq a)$.}

 {For the reported experiments, we use the prediction horizon $\Delta=12$ hours to compute the alerting policy. However, we note that different choices of $\Delta$, as seen in Eq. (\ref{landmark_failure_prob}), only change the scale of event probabilities; they do not affect the ordering of the patients and as a result, the choice of $\Delta$ does not affect the computation of any of the performance metrics reported in this paper. We also sweep the cost terms $L_1$, $L_2$, and $q$ (for the robust policy) to plot the TPR vs. FPR and TPR vs. PPV curves. To determine statistical significance of the results, we perform non-parametric bootstrap on the test set with bootstrap sample size 10 and report the average and standard error of the performance criteria.}

\textbf{Setup of the learning and inference algorithm:}  {We set the learning rate and maximum number of iterations for the global optimization, respectively, to $0.025$ and $1500$, set the mini-batch size to 2, and the number of Monte Carlo samples for reparameterization trick to 1000. We use L-BFGS-B \cite{byrd1995limited} for the local optimization with maximum number of iterations $500$. We set the number of inducing points ($M$) to 20 and the number of shared latent functions ($R$) to 2. These were set based on based visual analysis of the convergence results for the global parameters on the training data.}

\textbf{Implementation details:}  {We implemented the proposed model using TensorFlow \cite{Abadi2016} and GPflow \cite{gpflow} which automatically compute gradients of the ELBO with respect to all variables. The experiments reported in this section are obtained using the TensorFlow implementation running on a single machine with a 4-core 2.8 GHz CPU and 64 GB RAM. Local optimization of the parameters for an individual within each iteration of the learning algorithm takes on average $3$ seconds. This step is the main computational bottleneck; however, it is embarrassingly parallelizable and a distributed version of the algorithm enables scaling to larger datasets with more patients and longitudinal signals.}

\subsection{ {Results}}
\subsubsection{Qualitative analysis of example patients:}

First, we qualitatively investigate the ability of the proposed model---from hereon referred to as J-LTM---to model the longitudinal data and estimate the event probability. In Fig. \ref{sample_long}, we show the fit achieved by J-LTM on 10 longitudinal signals for two patients: a patient with septic shock (patient A) and a patient who did not experience shock (patient B). Despite the complexity of their physiologic data, J-LTM can fit the data well.  {We also see that J-LTM is robust against outliers; see, e.g., respiratory rate for patient B.}

 {Fig. \ref{sample_long} also shows the event probability computed for the following 40 hours conditioned on the data observed for each patient. J-LTM detects patient A as being at high risk on day 3.5 of his stay. The septic shock for this patient occurs 36 hours later. As shown in Fig. \ref{sample_long}, J-LTM computes a very high event probability with high confidence for patient A at prediction horizon $\Delta=36$ (the onset time of septic shock). In contrast, the event probability predicted for patient B, who did not have septic shock, is relatively low.
We can also gain insight about the main contributing factors for J-LTM's predictions by comparing different components of the weighted sums $\bfgamma^T\bfx_{it}$ and $\bfalpha^T\bar{\bff}_{it}$ in the hazard function. For instance, the top three factors for patient A are low GCS ($\alpha_{\text{gcs}}f_{\text{gcs}} = 0.27$), low PaO2 ($0.19$), and high heart rate ($0.13$). These are all clinically relevant factors which could contribute to organ failure and septic shock.}

\subsubsection{Interpreting model parameters:}

\textbf{Shared vs. signal-specific kernels:}  {In Fig. \ref{sample_long}, we see that HR, SBP, urine output, and respiratory rate (RR) are more densely sampled compared to other signals. In sparsely sampled signals, we expect that the shared latent components contribute more to the fit than the signal-specific kernels. To test this hypothesis, we compare the ratio of the weights of the shared and signal-specific kernels in Eq. (\ref{model}) ($||w_{id}||/||\kappa_{id}||, \forall i,d,$) across different signals. The median (and interquartile range (IQR)) of this ratio across all patients is $1.98$ ($9.06$) for HR, $8.94$ ($52.00$) for SBP, $118.89$ ($855.34$) for PaO2, and $281.02$ ($4289.40$) for WBC. We see that the coefficients of the shared functions are much greater than the weight of signal-specific kernels for sparse signals such as PaO2 and WBC.}

\textbf{Capturing correlations across signals:}  {The shared latent functions also help J-LTM capture correlations across signals. To evaluate the correlation patterns discovered by J-LTM, we compute the correlation coefficient between $w_{id1}$ and $w_{id2}$ across different signals of all patients (J-LTM has two shared latent functions; $R=2$). Some signals with highest cross-correlations are RR and HR with correlation coefficient $0.40$, urine output and SBP, $0.21$, and creatinine and BUN, $0.17$. These signals are in fact known to be related to each other. For example, creatinine and BUN are both measures of kidney function which are typically correlated.}

\subsubsection{Quantitative evaluation:}
\begin{figure}[t]
\centering
\begin{subfigure}{0.24\textwidth}
\centering
\vspace{-1mm}\includegraphics[scale=1]{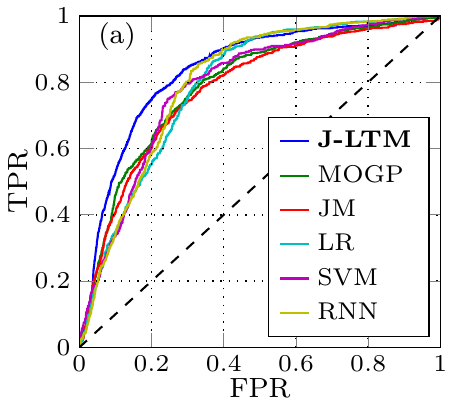}
\end{subfigure}\hspace{1mm}\hfill
\begin{subfigure}{0.24\textwidth}
\centering
\vspace{-1mm}\includegraphics[scale=1]{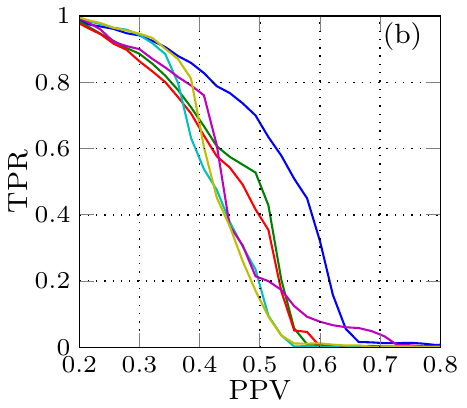}
\end{subfigure}\hfill
\vspace{0mm}
\begin{subfigure}{0.24\textwidth}
\centering
\includegraphics[scale=1]{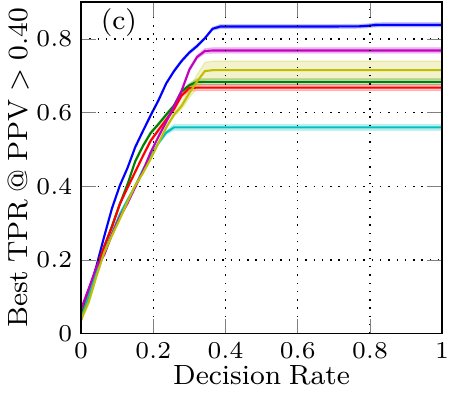}
\end{subfigure}\hfill\vspace{-2mm}
\begin{subfigure}{0.24\textwidth}
\centering
\includegraphics[scale=1]{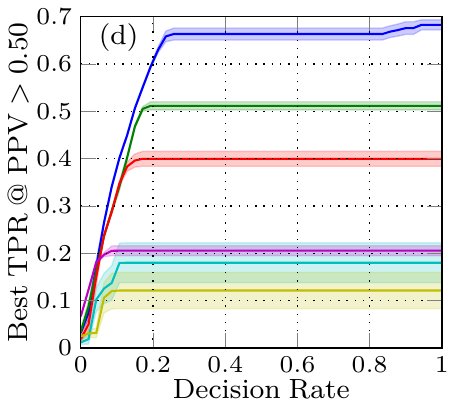}
\end{subfigure}\hfill\vspace{-0mm}
\caption{(a) ROC curves. (b) Maximum TPR obtained at each FAR level. (c) and (d) the best TPR achieved at any decision rate fixing PPV$>0.4$ and PPV$>0.5$, respectively.}\vspace{0mm}\label{roc_fig}
\end{figure}

\textbf{TPR vs. FPR:} Next, we quantitatively evaluate performance of J-LTM. We report the ROC curves (TPR vs. FPR) for J-LTM and the baseline methods (MoGP, JM, LR, SVM, and RNN) in Fig. \ref{roc_fig}a. To plot the ROC curve for each method, we performed grid search on the relative cost terms $L_1$ and $L_2$ and $q$ (for the robust policy), and recorded the obtained FPR and TPR pairs. J-LTM achieves an AUC (std. error) of $0.84$ ($0.005$) and outperforms MoGP, JM, LR, SVM, and RNN with AUCs $0.79$ ($0.006$), $0.78$ ($0.008$), $0.80$ ($0.005$), $0.79$ ($0.007$), and $0.80$ ($0.006$), respectively. As shown in Fig. \ref{roc_fig}a, the increased TPR for J-LTM compared to the baseline methods primarily occurs for FPRs ranging from $0.1-0.4$, the range most relevant for practical use. In particular, at FPR $=0.2$, true positive rate for J-LTM is $0.75$ (std. error $0.004$). At the same FPR, TPR for MoGP, JM, LR, SVM, and RNN are, respectively, $0.62$ ($0.006$), $0.61$ ($0.006$), $0.57$ ($0.003$), $0.62$ ($0.006$), and $0.59$ ($0.004$).

\textbf{TPR vs. PPV:} Fig. \ref{roc_fig}a compares performance using the TPR and FPR but does not make explicit the number of true alerts. An important performance criterion for alerting systems is positive predictive probability (PPV), the ratio of true positives to the total number of alarms. Every positive prediction by the classifier requires attendance and investigation by the clinicians. Therefore, a low PPV rate increases the workload of the clinicians and causes alarm fatigue. An ideal classifier detects patients with septic shock (high TPR) with low false alarms (high PPV). In Fig. \ref{roc_fig}b, we plot the maximum TPR obtained at each PPV level for J-LTM and the baselines.  {We sweep $L_1$, $L_2$, and $q$ (for the robust policy) and recorded the best TPR achieved at each PPV level.} We can see that at any TPR, the PPV for J-LTM is greater than that of all baselines. In particular, in the range of TPR from $0.4$-$0.6$, J-LTM shows $13\%$-$23\%$ improvement in PPV over MoGP, the next best baseline, and $18\%$-$26\%$ improvement in PPV over JM and $31\%$-$36\%$ over LR, methods typically implemented in standard-of-care tools. From a practical standpoint, each evaluation leads to a context switch and can cost the caregiver $30$-$40$ minutes; a \textit{$18\%$-$36\%$ improvement in the PPV can amount to many hours saved daily}.

To elaborate on this comparison further, we report TPR and PPV for each method as a function of the number of decisions made (i.e., at $1$, all models choose to make a decision for every instance). At a given decision rate, each model may abstain on a different subset of patients. In Fig. \ref{roc_fig}c and \ref{roc_fig}d, we show the best TPR achieved at any given decision rate for two different settings of the minimum PPV. In Fig. \ref{roc_fig}c, for example, at every abstention rate, we plot the best TPR achieved for every model with the PPV of greater than $40\%$. J-LTM achieves significantly higher TPR than baseline methods at all decision rates. In other words, at any given decision rate, J-LTM is able to more correctly identify the subset of instances on whom it can make predictions. Similar plots are shown in Fig. \ref{roc_fig}d: the maximum TPR with PPV$>$0.5 for J-LTM over all decision rates is $0.68$ (std. error $0.01$). This is significantly greater than the best TPR at the same PPV level for MoGP, $0.51$ ($0.008$), JM, $0.40$ ($0.02$), LR, $0.18$ ($0.04$), SVM, $0.21$ ($0.01$), and RNN, $0.12$ ($0.038$). A natural question to ask is whether the reported TPRs are good enough for practical use. The best standard-of-care tools implement the LR or JM baselines without abstention. This corresponds to the performance of these methods in Figs. \ref{roc_fig}c and \ref{roc_fig}d at the decision rate of $1$. As shown, the gain in TPR achieved by J-LTM are \emph{large} for both PPV settings.

\vspace{-3mm}
\section{Conclusion}\label{conc_section}

We propose a probabilistic framework for improving reliability of event prediction by incorporating uncertainty due to missingness in the longitudinal data. The proposed approach comprised several key innovations. First, we developed a flexible Bayesian nonparametric model for jointly modeling high-dimensional, continuous-valued longitudinal and event time data. In order to facilitate scaling to large datasets, we proposed a stochastic variational inference algorithm that leveraged sparse-GP techniques; this significantly reduced complexity of inference for joint-modeling from  {cubic in the number of signals ($D$) and the number of measurements per signal ($N$) to linear in both $D$ and $N$.} Compared to state-of-the-art in joint modeling, our approach scales to datasets that are several order of magnitude larger without compromising on model expressiveness. Our use of a joint-model enabled computation of the event probabilities conditioned on irregularly sampled longitudinal data. Second, we derived a policy for event prediction that incorporates the uncertainty associated with the event probability to abstain from making decisions when the alert is likely to be incorrect. On an important and challenging task of predicting impending in-hospital adverse events, we demonstrated that the proposed model can scale to time-series with many measurements per patient, estimate good fits, and significantly improve event prediction performance over state-of-the-art alternatives.

\ifCLASSOPTIONcompsoc
  \section*{Acknowledgments}
\else
  \section*{Acknowledgment}
\fi
The authors thank Wenbo Pan for his help with an initial implementation of the algorithm and Katharine Henry and Andong Zhang for their help with data.

\vspace{-2mm}

\balance

\begin{IEEEbiography}[{\includegraphics[width=1in,height=1.25in,clip,keepaspectratio]{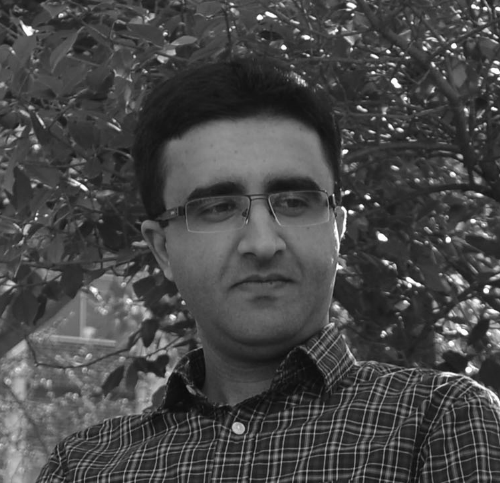}}]{Hossein Soleimani} received the PhD degree from Pennsylvania State University, PA, in 2016, and the MSc degree from University of Tehran, Tehran, Iran, in 2011, and the BSc degree from Ferdowsi University of Mashhad, Mashhad, Iran, in 2008, all in electrical engineering. He is currently a postdoctoral fellow in Johns Hopkins University, Baltimore, MD. His research interests include machine learning, healthcare, probabilistic graphical models, approximate posterior inference, and statistical modeling.
\end{IEEEbiography}

\begin{IEEEbiography}[{\includegraphics[width=1in,height=1.25in,clip,keepaspectratio]{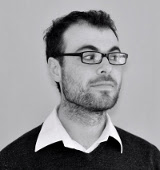}}]{James Hensman}
received the MEng and PhD degrees both in mechanical engineering from the University of Sheffield in 2005 and 2009, respectively. Following a Doctoral-Prize Fellowship in 2010, he joined Professors Rattray and Lawrence as a postdoc in machine learning for computational biology. His research interests include approximate Bayesian inference for large and complex systems in biology and biomedicine. He has now been awarded a Career Development Fellowship from the MRC in biostatistics to study inference methods in high-throughput data
\end{IEEEbiography}

\begin{IEEEbiography}
[{\includegraphics[width=1in,height=1.25in,clip,keepaspectratio]{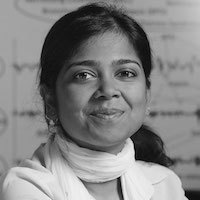}}]{Suchi Saria}
is a professor of computer science with joint appointments in Applied Mathematics \& Statistics and Health Policy at Johns Hopkins university. She is also the technical director for the Malone Center for Engineering in Healthcare at Johns Hopkins. Her research interests include statistical machine learning and decision-making under uncertainty. Prior to this, she received her PhD in computer science from Stanford University. 
\end{IEEEbiography}
  \newpage

\appendices
\section{}\label{appendA}
\subsection{Lemma}
Let $f(t)$ be a Gaussian process with mean $\mu(t)$ and kernel function $K(t, t')$. Then, $\int_0^T \rho(t) f(t) \dee t$ is Gaussian random variable with mean $\int_0^T \rho(t)\mu(t)\dee t$ and variance $\int_0^T \int_0^T \rho(t)K(t,t')\rho(t')\dee t \dee t'$.

\noindent \textbf{Proof:} We first note that $\bar{f}(T)=\int_0^T \rho(t) f(t) \dee t$ is a Riemann integral which can be approximated by $\bar{f}_n(T) = \frac{1}{n}\sum_{l=1}^{n}\rho(\frac{l}{n}T)f(\frac{l}{n}T)$. Clearly, $\bar{f}_n(T)\rightarrow \bar{f}(T)$ as $n \rightarrow \infty$ since $f(t)$ is a continuous function of time. Note that the random variables $f(\frac{l}{n}T), \forall l=1,2,...,n$ are correlated.

We compute the characteristic function of $\bar{f}_n(T)$, $M_n(\gamma)$:
\begin{align}
M_n(\gamma) &= E\exp\big(i \gamma \bar{f}_n(T)\big)= E\exp\big(i \gamma \frac{1}{n}\sum_{l=1}^{n}\rho(\frac{l}{n}T)f(\frac{l}{n}T)\big) \nonumber \\
&= \exp{\big(\frac{1}{n} \tilde{\gamma}^T \mu_n + \frac{1}{2}\frac{1}{n^2} \tilde{\gamma}^T K_{nn} \tilde{\gamma}\big)} \nonumber\\
&= \exp\big( \gamma \frac{1}{n}\sum_{l=1}^n \rho(\frac{l}{n}T)\mu(\frac{l}{n}T) \nonumber\\
&~+ \gamma^2\frac{1}{2}\frac{1}{n^2} \sum_{l=1}^{n}\sum_{l'=1}^{n}\rho(\frac{l}{n}T)K(\frac{l}{n}T, \frac{l'}{n}T)\rho(\frac{l'}{n}T)\big),
\end{align}
where in the third line we used the fact that $[f(\frac{1}{n}T), ..., f(\frac{n}{n}T)]^T = \mathcal{N}(\mu_n, K_{nn})$, with $\mu_n=[\mu(\frac{1}{n}T), ..., \mu(\frac{n}{n}T)]^T$ and $K_{nn} = K([\frac{1}{n}T,...,\frac{n}{n}T]^T, [\frac{1}{n}T,...,\frac{n}{n}T]^T)$. Also, we define $\tilde{\gamma} = [\gamma\rho(\frac{1}{n}T), ..., \gamma\rho(\frac{n}{n}T)]^T$.
Clearly, we have
\begin{align}
\lim_{n\rightarrow \infty} &M_n(\bar{f}_n(T)) = \exp\bigg( \gamma\int_0^T\rho(t)\mu(d)\dee t \nonumber \\
&~~~~~~~~~~~+ \frac{1}{2}\gamma^2\int_0^T\int_0^T \rho(t)K(t,t')\rho(t') \dee t\dee t'\bigg)\, ,  \label{cgf_normal}
\end{align}
which is the characteristic function of a Gaussian random variable with mean $\int_0^T\rho(t)\mu(d)\dee t$ and variance $\int_0^T\int_0^T \rho(t)K(t,t')\rho(t') \dee t\dee t'$.

Finally, due to continuity property of characteristic functions (see, e.g., \citet{billingsley2008probability}), we conclude that (\ref{cgf_normal}) is indeed  the characteristic function of the random variable $\bar{f}(T)$. This proves the claim. \QEDA

\subsection{Computing $E_{q(\bff)}\log p(\mathbf{T}, \delta|\bff(t))$}
We first compute the integral of one of the latent functions using the lemma proved above.

\subsubsection{Computing $\int_0^t \rho_c(t';t) g_r(t')\dee t'$:} Recall from section \ref{inf_sec} that the the variational approximation for $g_r$ is $q(\mathbf g_r) = \mathcal{GP}(\bfmu_{g_r}, \bfSigma_{g_r}), \forall r=1,...,R$, where $\bfmu_{g_r} = \bfK^{(r)}_{\bfN \bfZ}\bfK^{{(r)}^{-1}}_{\bfZ\bfZ}\bfm_r$ and $\bfSigma_{g_r} = \bfK^{(r)}_{\bfN\bfN}-\bfK^{(r)}_{\bfN\bfZ}\bfK^{{(r)}^{-1}}_{\bfZ\bfZ}(\mathbf I - \bfS_r\bfK^{{(r)}^{-1}}_{\bfZ\bfZ})\bfK^{(r)}_{\bfZ\bfN}$, with $\bfK^{(r)}_{\bfN\bfZ} = K_r(\mathbf{t}, \bfZ)$.

Using the lemma, we can easily show that the distribution of $\int_0^t \rho_c(t';t) g_r(t')\dee t'$ is $\mathcal{N}(\mu_{g_r}^{(t)}, \sigma^{2^{(t)}}_{g_r})$, where
\begin{align}
\mu_{g_r}^{(t)} &= \bar{K}^{(r)}_{t\bfZ}\bfK^{(r)^{-1}}_{\bfZ\bfZ}\bfm_r\, ,\nonumber\\
\sigma^{2^{(t)}}_{g_r} &= I^{g_r}_{t} - \bar{K}^{(r)}_{t\bfZ}\bfK^{{(r)}^{-1}}_{\bfZ\bfZ}(\mathbf I - \bfS_r\bfK^{{(r)}^{-1}}_{\bfZ\bfZ})\bar{K}^{(r)}_{\bfZ t}\, ,
\end{align}
with
\begin{align}
&\bar{K}^{(r)}_{t\bfZ}(z) = \frac{c'}{c+\frac{1}{2l_{g_r}}}\big[\exp\big(c(z-t)\big)-K_r(0, z)\big]\mathds{1}(0\leq z \leq t) \nonumber\\
&~~~~~~+ \frac{c'}{c-\frac{1}{2l_{g_r}}}\big[K_r(t, z)-\exp\big(c(z-t)\big)\big]\mathds{1}(t \leq z)\, ,\forall z \in \bfZ\, , \nonumber \\
I^{g_r}_{t} &= \bigg[1 + \exp{(-2ct)} - \frac{1}{2cl_{g_r}}(1 - \exp{(-2ct)})\nonumber\\
&~~~~~-2\exp\big(-(c+\frac{1}{2l_{g_r}})t\big) \bigg]\times \frac{{c'}^2}{c^2-\frac{1}{4l^2_{g_r}}}\, , \nonumber \\
c' &= \frac{c}{1 - \exp(-ct)}\, .
\end{align}
Here, $l_{g_r}$ is the length-scale of the kernel $k_{g_r}$. We similarly compute the variational distribution of the integrals of the signal-specific latent functions: $\int_0^t \rho_c(t';t) v_d(t')\dee t'$ $\sim$ $\mathcal{N}(\mu_{v_d}^{(t)}, \sigma^{2^{(t)}}_{v_d})$.

Using these, we compute $\bfalpha^T\int_0^t\rho_c(t';t)\bff_i (t')\dee t'\sim \mathcal{N}(\mu^{(t)}, \sigma^{2^{(t)}})$, where
$\mu^{(t)} = \sum_{d=1}^D \kappa'_d \mu_{v_d}^{(t)} + \sum_{r=1}^R \omega'_r \mu_{g_r}^{(t)}$, $\sigma^{2^{(t)}} = \sum_{d=1}^D {\kappa'}^2_d \sigma_{v_d}^{{(t)}^2} + \sum_{r=1}^R \omega'^2_r \sigma^{2^{(t)}}_{g_r}$,
with $\kappa'_d = \kappa_d \alpha_d$ and $\omega'_r = \sum_{d=1}^D \omega_{dr}\alpha_d$.

\subsubsection{Computing $E_{q(\bff)}\log p(\mathbf{T}, \delta|\bff)$:}
We compute
\begin{align}
&E_{q(\bff)}\log p(\mathbf{T},\delta|\bff(t)) = E_{q(\bff(t))}\Big[\log S(T_{l}|\bff,t) \nonumber \\
&+\mathds{1}(\delta=0)\log\lambda(T;t)+ \mathds{1}(\delta=2) \log F(\Delta T|\bff, t) \Big]\, ,
\label{t2e_elbo_terms}
\end{align}
where we replaced $p(\mathbf{T},\delta|\bff)$ as defined in (\ref{t2e_lkh}), factored out $S(T_{l}|\bff,t)$, and followed the dynamic approach for defining the hazard function described in section \ref{t2e_submodel}. Here,
\begin{align}
F(\Delta T|\bff, t) &= 1-S(T_r|\bff,t)/S(T_l|\bff,t)\nonumber\\
&=1-\exp\big(-\frac{1}{a} \lambda(T_{l};t)(\exp(a\Delta T)-1) \big)\, .
\end{align}
We also assumed $T_{l}=T$ when $\delta = 0$, and defined $\Delta T = T_{r}-T_{l}$.
The first two terms in (\ref{t2e_elbo_terms}) are computed analytically:
\begin{align}
&E_{q(\bff)}\log S(T_{l}|\bff,t) = \frac{-1}{a}(1 - \mathrm{e}^{-a(T_{l}-t)})E_{q(\bff)}\lambda(T_l;t)\, ,\nonumber\\
&E_{q(\bff)}\log\lambda(T;t) = (b+a(T_{l}-t) + \bfgamma^T\bfx_t+ \mu^{(t)})\, ,
\end{align}
where $E_{q(\bff)}\lambda(T_l;t) = \exp{\big(b+a(T_{l}-t) +\bfgamma^T\bfx_t+ \mu^{(t)} + \frac{1}{2}\sigma^{2^{(t)}}\big)}$.

The term related to interval censoring in (\ref{t2e_elbo_terms}) cannot be computed analytically. We also need to take derivative of this term with respect to parameters of the variational distribution $q(\bff)$ and time-to-event parameters. To do this, we use reparameterization tricks and compute Monte Carlo (MC) estimate of the expectation and the gradients \cite{Kingma2013}:
\begin{align}
E_{q(\bff(t))}\log F(\Delta T|\bff, t)\approx \frac{1}{N_0} \sum_{n=1}^{N_0} \log \tilde{F}_{\epsilon_n}(\Delta T|\bff, t)\, ,
\end{align}
where $\tilde{F}_{\epsilon_n}$ is computed using the hazard rate $\hat{\lambda}(T_{l};t,\epsilon_n) = \exp\big(b+a(T_{l}-t) +\bfgamma^T\bfx_t+ \mu^{(t)}+\sigma^{(t)}\epsilon_n\big)$. Here, $\epsilon_n\sim \mathcal{N}(0, 1)$, and $N_0$ is the MC sample size.

\section{}\label{appendB}
\subsection{Distribution of $H$}
Recall that $H\triangleq H(\Delta|\bar{f},t)$ depends $\bar{f}$ which, as described in section \ref{methods_section}, is itself a Gaussian random variable; $\bar{f}(t)\sim \mathcal{N}(\mu^{(t)}, \sigma^{2^{(t)}})$. Thus, $H$ is also a random variable whose distribution is computed based on the distribution of $\bar{f}$:

\begin{align}
p_H(h) = \frac{\mathcal{N}\bigg(\log\big(\frac{1}{k}\log (1-h)\big);b+\bfgamma^T\bfx_t+\mu^{(t)},\sigma^{2^{(t)}}\bigg)}{(h-1)\log(1-h)}\, ,
\label{failure_distn}
\end{align}
where $k \triangleq \frac{1}{a}(1-\exp(a\Delta))$.
As $\Delta \rightarrow \infty$, $p_H(h)$ converges to a degenerate distribution $p_H(h)=\mathds{1}(h=1)$. Similarly, $\Delta \rightarrow 0$ yields $p_H(h)=\mathds{1}(h=0)$.

We also note that q-quantiles of the distribution (\ref{failure_distn}) can be easily computed using q-quantiles of Gaussian distribution. Specifically, q-quantile of (\ref{failure_distn}) is $h^{(q)} = 1 - \exp\big(k \exp(v^{(q)})\big)$, where $v^{(q)}$ is the q-quantile of a Gaussian distribution with mean $b+\bfgamma^T\bfx_t+\mu^{(t)}$ and variance $\sigma^{2^{(t)}}$.

\subsection{Lemma}
The q-quantile of the random variable $1-H$ is $1-h^{(1-q)}$, where $h^{(1-q)}$ is the (1-q)-quantile of the random variable $H$ and $H\in[0,1]$.

\noindent \textbf{Proof:} {Let $H_{0}=1-H$, and suppose (1-q)-quantile of $H$ is $h^{(1-q)}$. Observe that
\begin{align}
q \hspace{-0.5mm}= \hspace{-1mm}\int_{0}^{h_0^{(q)}} \hspace{-4mm}p_{_{H_0}}(h_0)\dee h_0 =\hspace{-1mm} \int_{1-h_0^{(q)}}^1 p_{_{H}}(h)\dee h = 1-\int_0^{1-h_0^{(q)}}\hspace{-3mm}p_{_H}(h)\dee h.\nonumber
\end{align}
Thus, $\int_0^{1-h_0^{(q)}}p_{_H}(h)\dee h = 1-q$, and we conclude that $h^{(1-q)}=1-h_0^{(q)}$.}

\end{document}